\documentclass[authoryear,preprint]{elsarticle}

\usepackage{array}
\usepackage{hyperref}
\usepackage{bookmark}
\usepackage{graphicx}
\usepackage[subrefformat=parens, labelformat=parens]{subfig} 
\usepackage{url}   
\usepackage{amsmath}
\usepackage[margin=3.5cm]{geometry}
\usepackage{gensymb}
\usepackage{verbatim}
\usepackage{natbib}

\usepackage[usenames, dvipsnames]{color}
\usepackage{setspace}

\journal{Journal of Biomechanics}
\usepackage{etoolbox}

\begin{document}
	
\begin{frontmatter}
% paper title
	\title{An ocular biomechanics environment for reinforcement learning}
	%\title{Learning to Look: Using Deep Reinforcement Learning and Ocular Biomechanics to Perform Fixations and Saccades}
   
    \author[iisri]{Julie~Iskander\corref{mycorr}}
    \cortext[mycorr]{Corresponding author}
    \ead{julie.iskander@research.deakin.edu.au}
    \author[iisri]{Mohammed~Hossny}
    
    \address[iisri] {Institute for Intelligent Systems Research and Innovation (IISRI)\\
    Deakin University, Australia}

    \begin{abstract}
       Reinforcement learning has been applied to human movement through physiologically-based biomechanical models to add insights into the neural control of these movements; it is also useful in the design of prosthetics and robotics. In this paper, we extend the use of reinforcement learning into controlling an ocular biomechanical system to perform saccades, which is one of the fastest eye movement systems. We describe an ocular environment and an agent trained using Deep Deterministic Policy Gradients method to perform saccades. The agent was able to match the desired eye position with a mean deviation angle of $3.5\degree~\pm 1.25\degree$. The proposed framework is a first step towards using the capabilities of deep reinforcement learning to enhance our understanding of ocular biomechanics. 
    \end{abstract}
    \smallbreak
    \smallbreak
\end{frontmatter}
\doublespacing

\section{Introduction}
	
	Eye movement is one of the most complex, and the fastest movement that our body performs~\citep{leigh2015neurology}; the different eye movement systems are tightly coupled with mental, cognitive and psychological states of the individual~\citep{wong2008eye,iskander2018eye}. One of the most studied eye movement systems is saccade, which shifts the gaze direction to a new point of interest rapidly~\citep{gilchrist2011saccades}. Simulating the neural control of the muscles that could efficiently achieve eye movement through biomechanical simulation and analysis is an essential tool for studying different eye movement systems in normal and pathological cases~\cite{iskander2018review,iskander2018ocular}.
	
\begin{figure*}[t]
	\centering
	%%%trim={<left> <lower> <right> <upper>}
	\subfloat[]{
	 \includegraphics[width=0.33\linewidth]{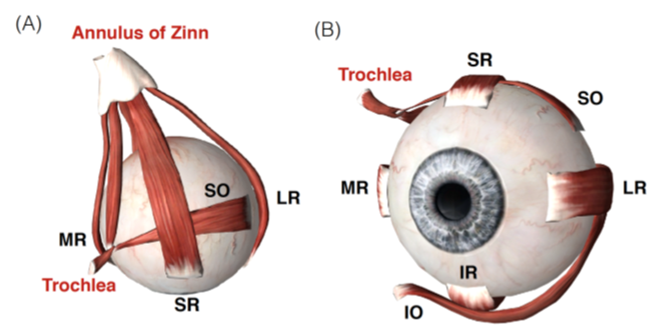}
		\label{fig:muscles}}
	\subfloat[]{
	 \includegraphics[width=0.3\linewidth,scale=0.25]{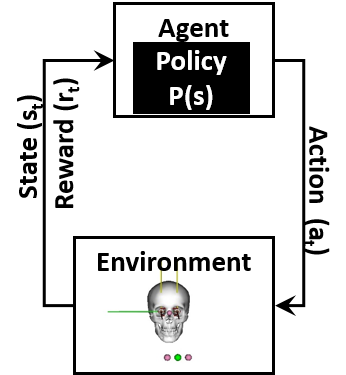}
		\label{fig:model}}
    \subfloat[]{
        \includegraphics[width=0.33\linewidth]{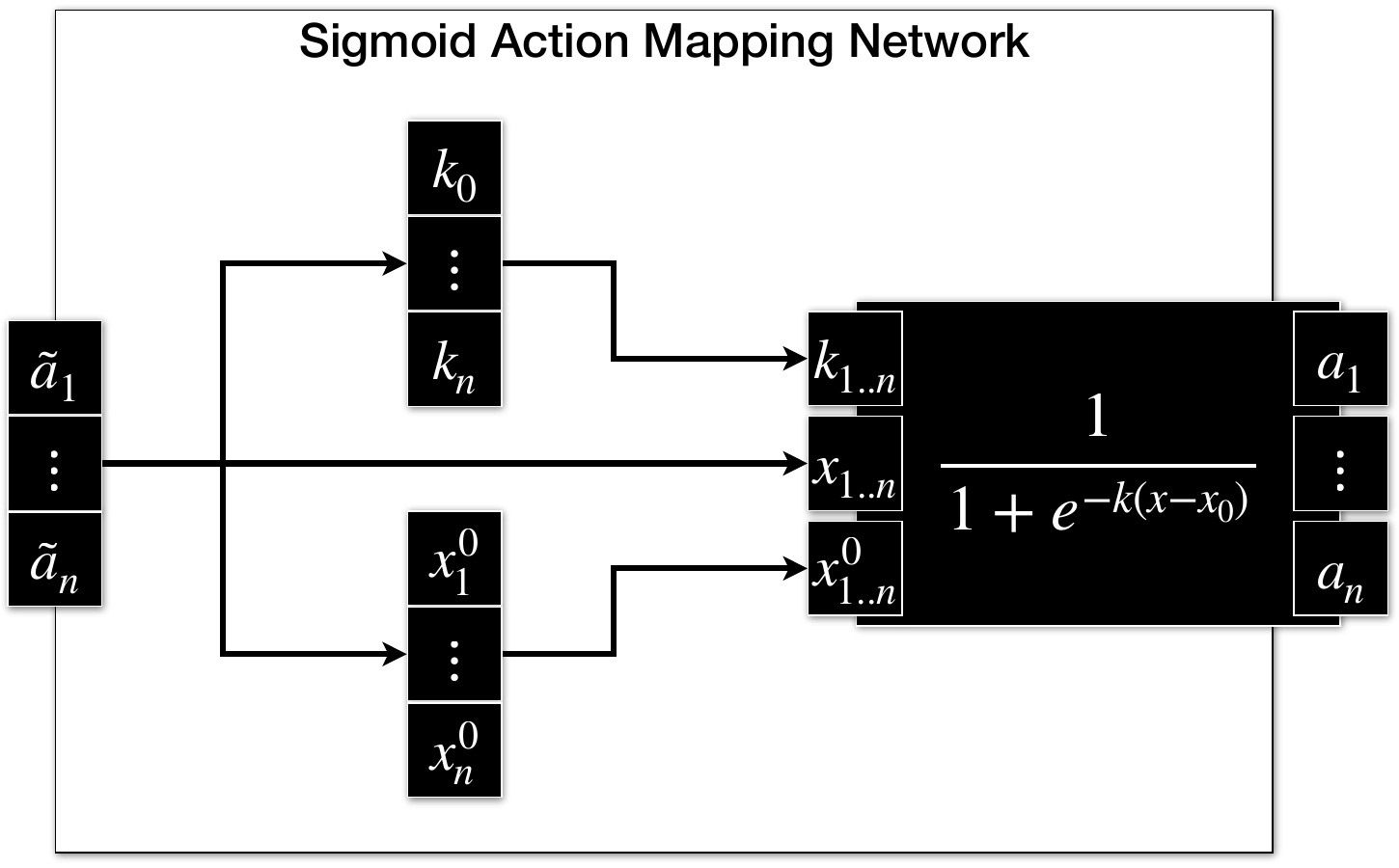}	
         \label{fig:actmap_sigmoid}}\\
	\subfloat[]{
		\includegraphics[width=0.95\linewidth]{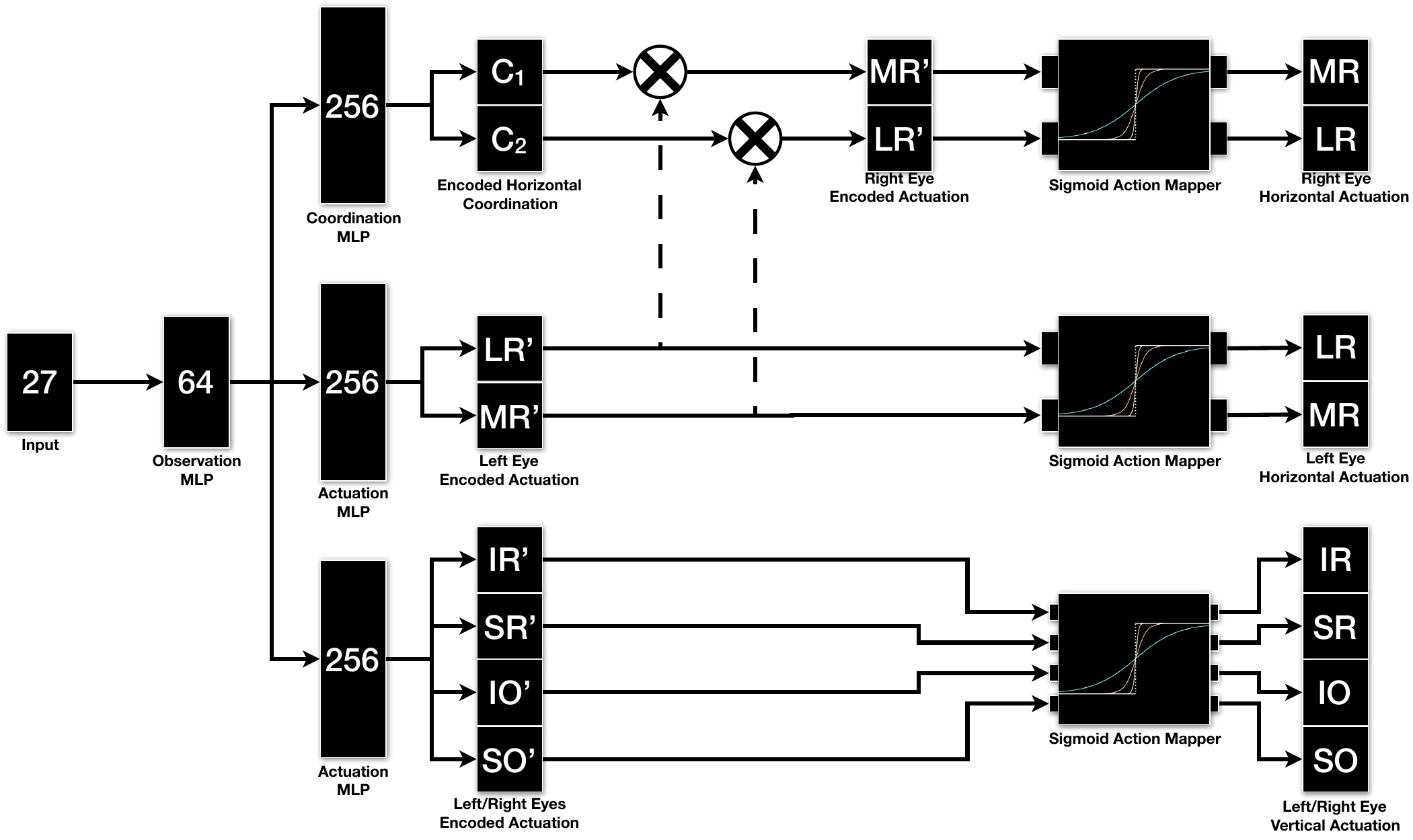}	
		\label{fig:eomctrl}}
	
	\caption{(a) Human eye with muscles shown from (A) Superior view, (B) Anterior view. The names and abbreviations of each muscle is shown in Table~\ref{tab:EOMfunc} along with their actions. From~\citep{iskander2018review} (b) Schematic describing how reinforcement learning works. The neuro-musculoskeletal model used is shown. The model is made up of a skull with two eye, each eye has six extraocular muscles attached to it. The green sphere presents the object of interest, the pink spheres shows the direction of gaze of the eye (PoG). (c) Sigmoid action mapping network used. (d) Policy learning network used}
	\label{fig:eye}
\end{figure*}

	The horizontal, vertical and torsional eye movements are created through the activation of six extraocular muscles (EOM), Fig.~\subref*{fig:muscles}~\cite{iskander2018review}. The action/name of each of the six EOM are described in Table~\ref{tab:EOMfunc}. To move the eye to the right, the right LR and the left MR are activated (agonists) while the right MR and left LR are inhibited (antagonists)~\citep{sherrington1893ii,hering1977theory}. The opposite happens for leftward eye movement. To move the eye vertically, the SR, IR, SO and IO muscles are activated/inhibited~\citep{purves2001neural,scudder2002brainstem,sparks2002brainstem}. Next, we will introduce the concept of reinforcement learning as it will be used to control an ocular biomechanical system.

	Reinforcement learning (RL) is a form of artificial intelligence that is different and fundamentally more difficult than supervised learning. According to Sutton and Barto~\citep{sutton2018introduction}, RL is learning how to map a situation into an action such that a numerical reward is maximised. In the case of biomechanical studies, the situations is the biomechanical model state ($s$), and the actions is the muscle excitation signals ($a$), where as the numerical reward ($r$) reflects the desired movement. And, thus the RL agent learns to map each state into an efficient action that maximises the reward. In addition, we have a policy ($P(s)$) which defines the strategy the agent uses, Fig.~\subref*{fig:model}. The policy $P(s)$ is the mapping from the $s$ to $a$. As the agent proceeds in the training phase, $P(s)$ evolves to produce the highest cumulative reward over time. The DRL training takes place in episodes. Each episode is a trial that allows the agent to explore the environment and to receive a reward based on how good or bad its behaviour was. The cumulative reward throughout an episode determines effectiveness of the training. RL has been used to model neural control of  movements such as walking, running and standing~\citep{HossnyIskander2020_dontfall,kidzinski2018learning,kidzinski2018learningsolu,kidzinski2020artificial}.

    \begin{table*}[b]
	\centering
	\caption[Actions of extraocular muscles.]{Action/Name of extraocular muscles. The table shows the  action direction of each muscle\citep{von2002binocular}. More details in~\citep{iskander2018review,leigh2015neurology,wong2008eye}}
        \begin{tabular}[width=\textwidth]{|l|ccc|}
        \hline
    		Muscle & Primary  & Secondary  & Tertiary \\
    		\hline
    		\hline
    		& ~& ~&\\
    		Lateral Rectus (LR) & Abduction & - &- \\
    		Medial Rectus (MR)  & Addution & - & - \\
    		Superior Rectus (SR) & Supraduction & Incycloduction & Adduction\\
    		Inferior Rectus  (IR)& Infraduction& Excycloduction& Adduction\\
    		Superior Oblique (SO) & Incycloduction & Infraduction & Abduction\\
    		Inferior Oblique  (IO) & Excycloduction & Supraduction & Abduction\\ 
    		\hline
    	\end{tabular}
		
	\label{tab:EOMfunc}
    \end{table*}
	
	In the following, we present an ocular biomechanics simulation environment suitable for RL; and we train an agent to control the ocular biomechanics system by producing adequate muscle excitation signals to perform saccades. 

\section{Methods}
	\label{sc:methods}
	The DRL environment used is made up of five components, as follows, the (1) neuro-musculoskeletal model;
	(2) the continuous state vector, that is produced from the environment; (3) the continuous action vector (muscle excitation signal), used to activate the neuro-musculoskeletal model; (4) the reward function to be maximise; and finally (5)  the training mechanism used, which includes the actor and critic neural networks used.
	    
    \subsection{Ocular Biomechanics Environment}
 The proposed DRL environment is based on OpenAI~\citep{brockman2016openai}, OpenSim~\citep{kidzinski2018learning,seth2018opensim} and ocular biomechanics~\citep{iskander2018ocular,iskander2019using,iskander2018biomechanical}. The neuro-musculoskeletal model, Fig.\ref{fig:model}, consist of a skull and two eyes; the skull has no degrees-of-freedom (DoFs). Each eye has six extraocular muscles which rotated the eye around three axis $x$, $y$, and $z$. The model uses Millard muscle model~\citep{millard2013flexing} for the muscles. 
     
     \subsection{The state vector, action vector and reward function}
     	The state($s$) vector includes 27 values, as follows:
     \begin{itemize}
         \item The 3D position of the object of interest;
         \item the 3D direction of gaze of each eye (point of gaze, POG);
         \item the 3D orientation of each eye; and
         \item the activation of the 12 muscles.
     \end{itemize}
     All measured values are in radians and meters.
     \smallbreak
     The action ($a$) includes 12 values, in the range [0,1]. They represent the 12 extraocular muscles excitation signals~\citep{millard2013flexing,thelen2003adjustment}. The step size of the environment is 0.01 s, i.e. a step every 10 ms. 
     \smallbreak
     The reward function is defined as:
    \begin{equation} \label{eq:rewfun}
    \begin{split}
        r_t=-w_1*\lVert dist_{RO}\rVert^2 - w_2*\lVert dist_{LO}\rVert^2  \\
            - w_3*\lVert dist_{LR}\rVert - w_4* {lr_y}^2 - w_5*{lr_z}),
    \end{split}
    \end{equation}
     
    \noindent where $dist_{RO}$ and $dist_{LO}$ is the distance between the object of interest and the direction of gaze of each eye (R and L), $dist_{LR}$ is the distance between the POG of the two eyes, ${lr_y}$ is the difference in the vertical position of both eyes, $lr_z$ is a binary value that indicated whether crossed eyes occurred or not. Crossed eyes is measured by the horizontal position of the eye POG, where the right eye POG should lie to the right of the left eye POG. Finally, $w_1$, $w_2$, $w_3$, $w_4$,  and $w_5$ are weights whose values are 16,16,32,64 and 64, respectively. The objective is that the reward should approach zero.
    
    \subsubsection{Training Methodology}
     The state and action vectors are continuous,  therefore Deep Deterministic Policy Gradients (DDPG) is used. DDPG is an off-policy, actor-critic algorithm for continuous observation and action spaces~\citep{lillicrap2015continuous}. Actor-critic based RL uses two modules, an actor and a critic. The actor learns a policy that maps the current state into an action, while the critic assesses the anticipated reward based on the current observation and the actor's action. DDPG, also, uses an experience replay buffer, to store previous experiences. The experience replay buffer is used randomly to train the actor and critic neural networks; that is why it is categorised as off-policy~\citep{konda2000actor,silver2014deterministic,lillicrap2015continuous}.
    Testing was done on two phases. First, we tested each milestone policy (actor network), by using it to run 10 episodes (100 step each). A video of the process is in the supplementary material. At the start of each episode, the target object of interest is located at ($x$, $y$, $z$) = (1,0,0), which is centrally in front of the two eyes. Then, the object is randomly moved by ($x$, $y$+$dy$, $z$+$dz$) where $dy$ and $dz$ range between [-0.16, 0.16] and [-0.32, 0.32], respectively.
    In the second testing phase, we used the policy (actor network) achieved in the final milestone. The test defined nine positions for the object of interest. The nine positions create a 3x3 grid with points at 0.1 m distance from the initial position in $\pm y$ and $\pm z$ directions. For each position, 50 episodes were performed, each containing 100 steps. 
\smallbreak

\subsubsection{RL Agent Network Structure}
    The actor and critic neural networks has 4 layers. All layers, except the final layer, utilise Rectified Linear Unit (ReLU) as the activation function. The final layer, in the actor, is an action mapping network, whereas the final layer, in the critic, receive no activation (Linear). 
    In order to provide fine tuning over the produced excitation signals, action mapping network is added to the actor, to infer the parameters of the sigmoid function ($k, x_0$) for each muscle independently~\citep{HossnyIskander2020_dontfall,HossnyEtal2020_PTANH}, Fig.\subref*{fig:actmap_sigmoid} and Fig.\subref*{fig:eomctrl}. The sigmoid activation function is, 
    \begin{equation} \label{eq:sigmoid}
        \frac{1}{1+e^{-k(x-x_0)}},
    \end{equation}
    \noindent where $k$ controls the steepness of the curve and $x_0$ controls the minimum value as dictated by the intercept with the y-axis. 

    The actor and the critic neural networks have separate Adam optimisers~\citep{kingma2014adam}. Training took 10000 episodes (100 step each). The batch size used is 64 and the learning rate of 0.001 is used.

    We adapted a model specific neural network architecture. This allowed us to enforce the LR/MR muscle coordination between the left and the right eyes, where the right LR and the left MR are innervated similarly and the same for the right MR and the left LR~\citep{wong2008eye,purves2001neural}. In addition, since the LR and MR muscles have different properties~\citep{iskander2018ocular}, we added a coordination neural network that infers two constants, $C_1$ and $C_2$, to allow for fine tuning between left LR and right MR as follows;
 \begin{eqnarray} \label{eq:coord}
        LR_r &=& C_1 \cdot MR_l,\\
        MR_r &=& C_2 \cdot LR_l.
\end{eqnarray}

    In the case of the SR, IR, SO and IO muscles, the left and the right eye used excitation signals inferred from the same neural network.

\section{Results}
	\label{sc:results}
\begin{figure*}
	\centering
	%%%trim={<left> <lower> <right> <upper>}
	\subfloat[]{
		\includegraphics[trim={0cm 0cm 0cm 0.5cm},clip,width=0.7\linewidth]{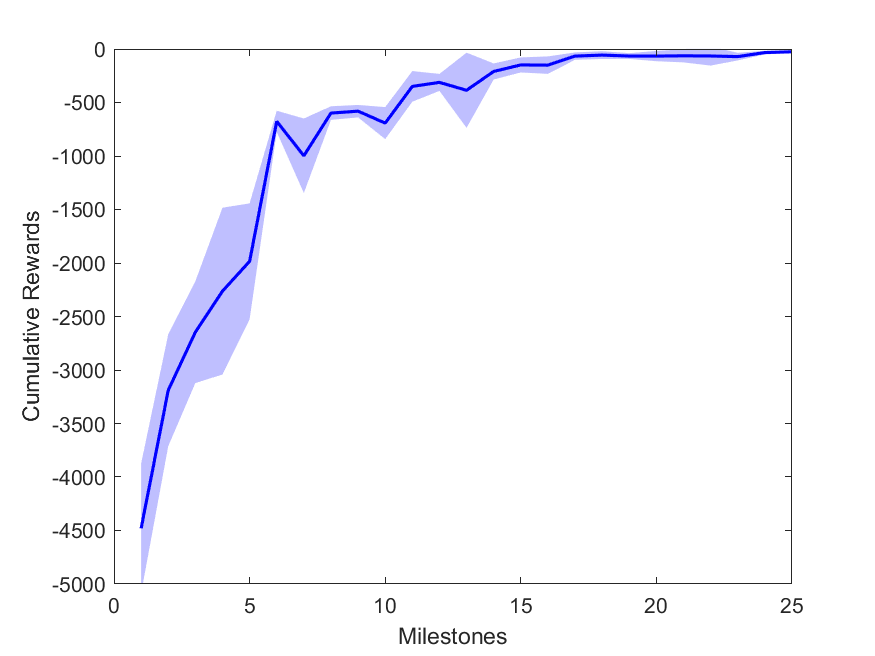}
		\label{fig:milestone_reward}}\\
	\subfloat[]{
		\includegraphics[trim={0.5cm 0cm 1cm 0cm},clip, width=0.4\linewidth]{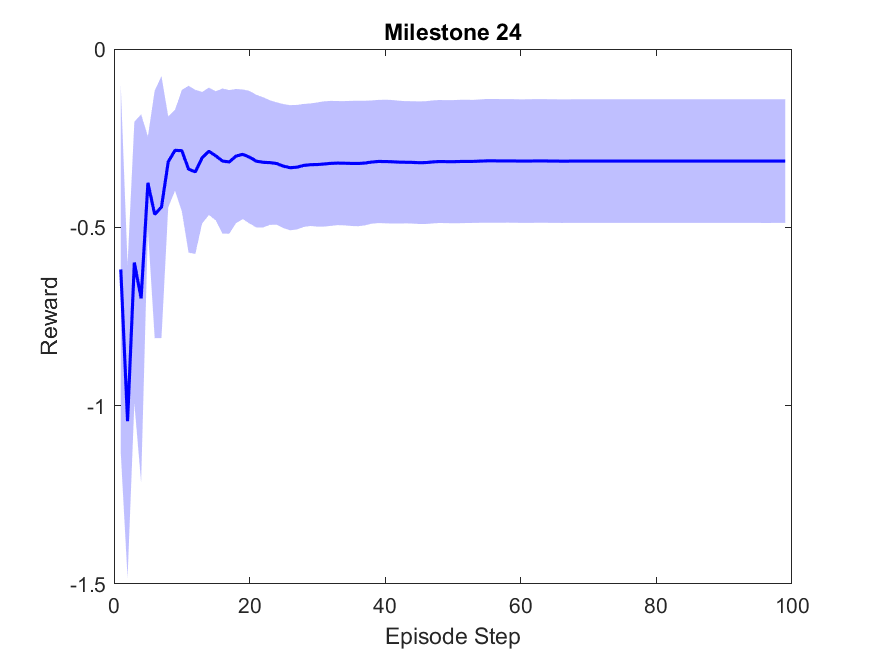}	
		\label{fig:M24}}
	\subfloat[]{
		\includegraphics[trim={0.5cm 0cm 1cm 0cm},clip,width=0.4\linewidth]{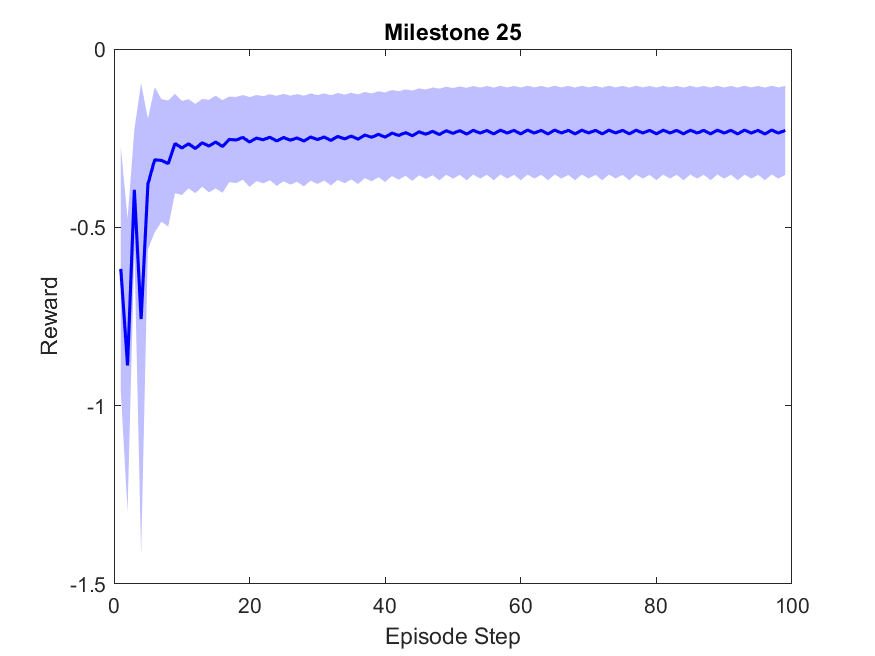}	
		\label{fig:M25}}\\
		\subfloat[]{
		\includegraphics[trim={0.5cm 0cm 1cm 0cm},clip,width=0.4\linewidth]{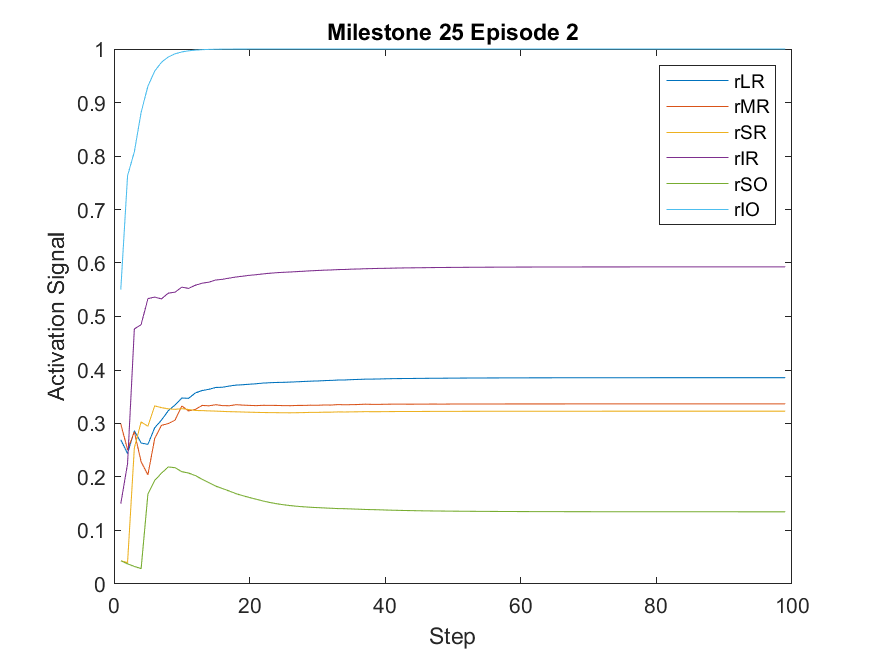}	
		\label{fig:RA}}
	\subfloat[]{
		\includegraphics[trim={0.5cm 0cm 1cm 0cm},clip,width=0.4\linewidth]{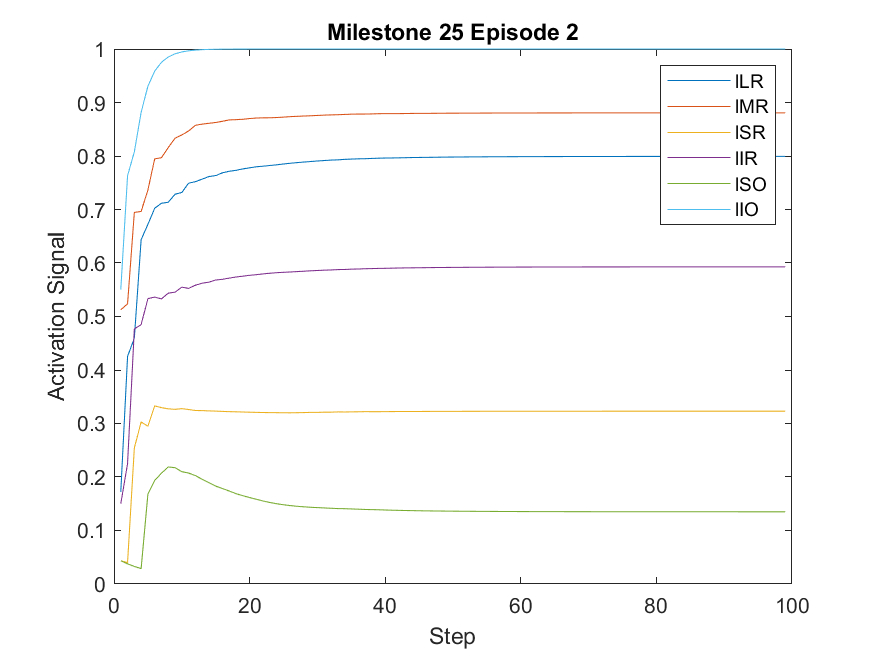}	
		\label{fig:RL}}\\
	
	\caption{Training and testing results. (a) Mean cumulative reward achieved at each milestone during training, the shaded part represents the standard deviation. (b) and (c) show the results of the first phase testing on policy produced from milestone 24 and milestone 25, respectively, (d) and (e) show the muscle activation signal produced during the testing phase one on Milestone 25 Episode 2 for the right and left eye, respectively.}
	\label{fig:results}
\end{figure*}

	During training, the agent achieved 26 milestones to reach the final trained state. Each milestone defined an increase in the cumulative reward and thus, a better policy ($P(s)$) was achieved, Fig. \subref*{fig:milestone_reward}.
	\smallbreak

	Figure \ref{fig:results} shows the results of the first testing phase done on each milestone. Figure \subref*{fig:milestone_reward} shows the cumulative reward achieved at each milestone. As the training proceeded, the cumulative reward improved and approached zero which is the optimal reward value. Figures \subref*{fig:M24} and \subref*{fig:M25} show the rewards achieved at each step for the last two milestones. Figures \subref*{fig:RA} and \subref*{fig:RL} shows the muscle activation signal of the right and left eye, respectively during the episode number 2 of the last milestone. The object of interest was displaced by $dy$=0.1231 and $dz$=-0.1112. For the eyes to follow this object, both eyes has to be elevated; the right eye has to be adducted; and the left eye abducted. Figures \subref*{fig:RA} and \subref*{fig:RL} shows high activation of IO which causes elevation and abduction of the eye in contrast to SO which has a decreasing activation as it is an agonist muscle.

\begin{figure*}
	\centering	
	%%%trim={<left> <lower> <right> <upper>}
	\subfloat[(-0.1,0.1)]{
		\includegraphics[trim={1.9cm 9.5cm 1.9cm 10cm},clip,width=0.33\linewidth]{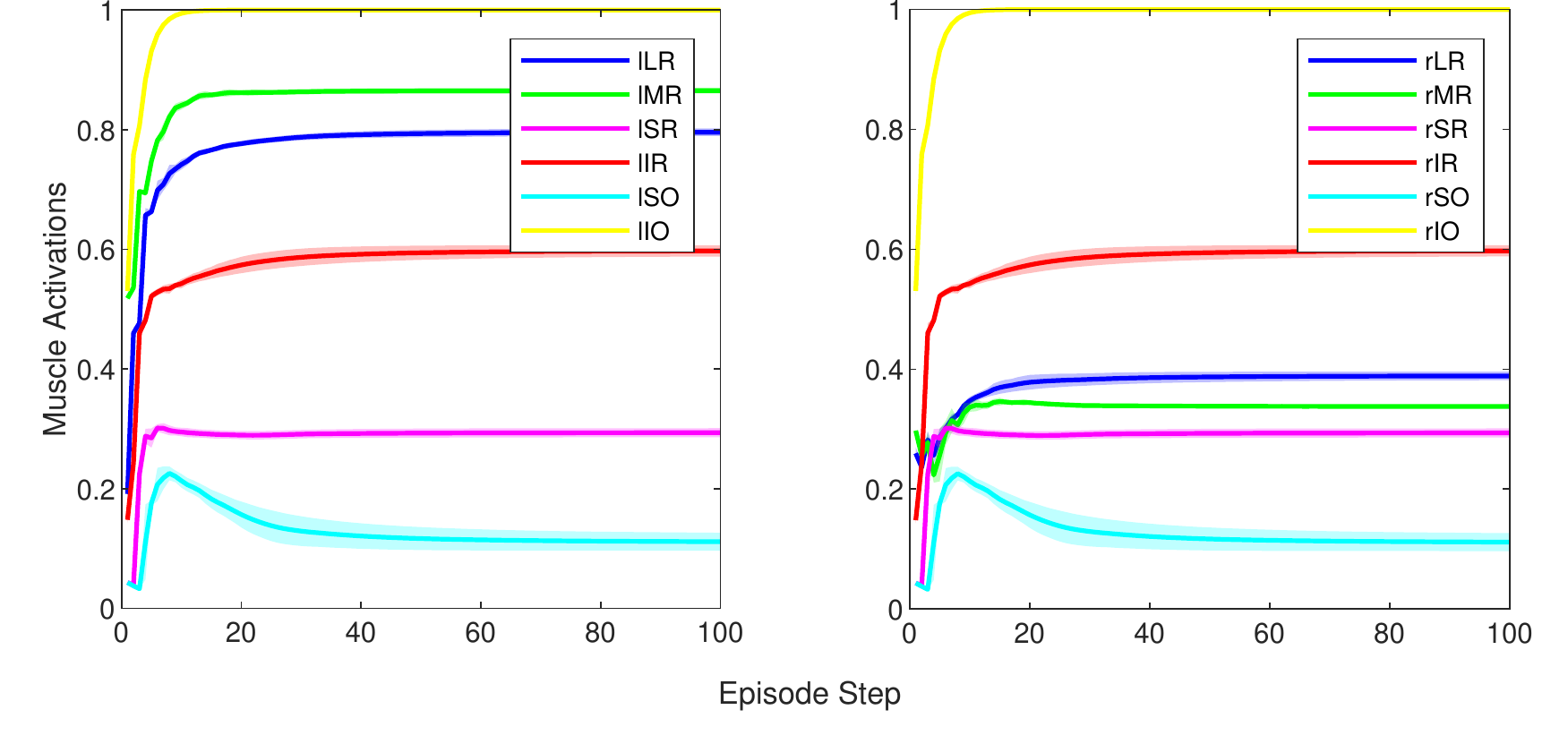}
		\label{fig:act-0101}}
	\subfloat[(0.0,0.1)]{
		\includegraphics[trim={1.9cm 9.5cm 1.9cm 10cm},clip,width=0.33\linewidth]{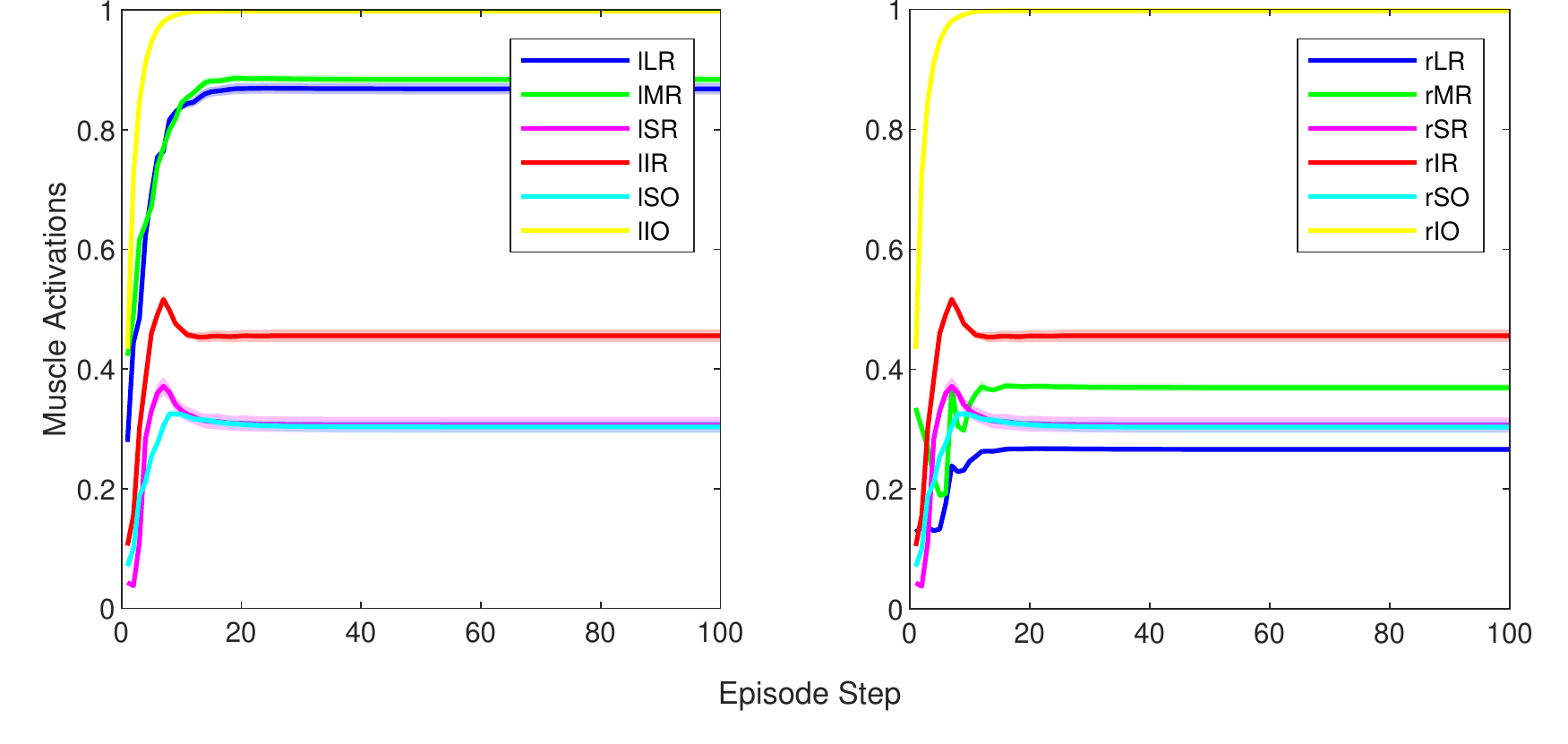}		
		\label{fig:act00.1}}
	\subfloat[(0.1,0.1)]{
		\includegraphics[trim={1.9cm 9.5cm 1.9cm 10cm},clip,width=0.33\linewidth]{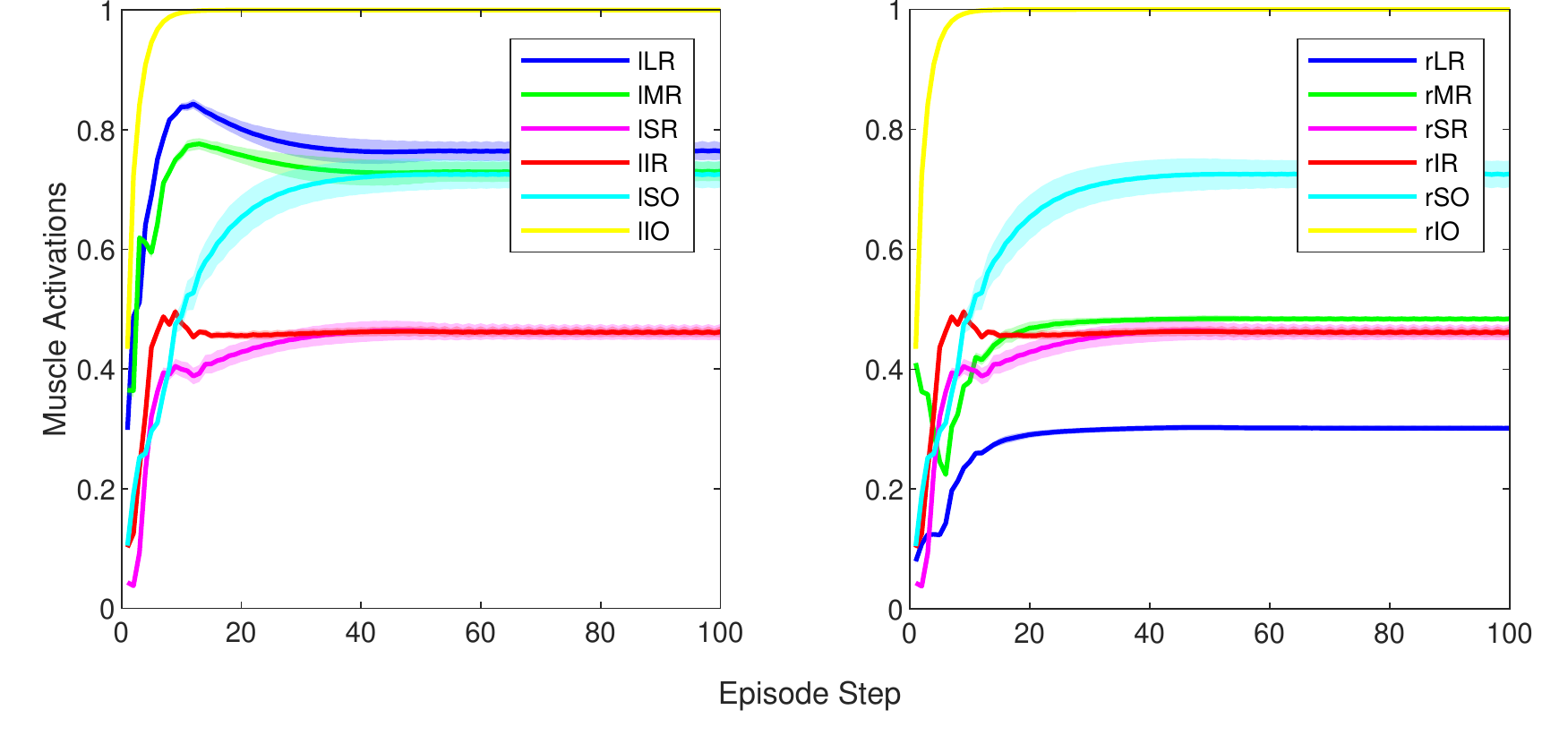}
		\label{fig:act0.10.1}}
		\\
	\subfloat[(-0.1,0.0)]{
			\includegraphics[trim={1.9cm 9.5cm 1.9cm 10cm},clip,width=0.33\linewidth]{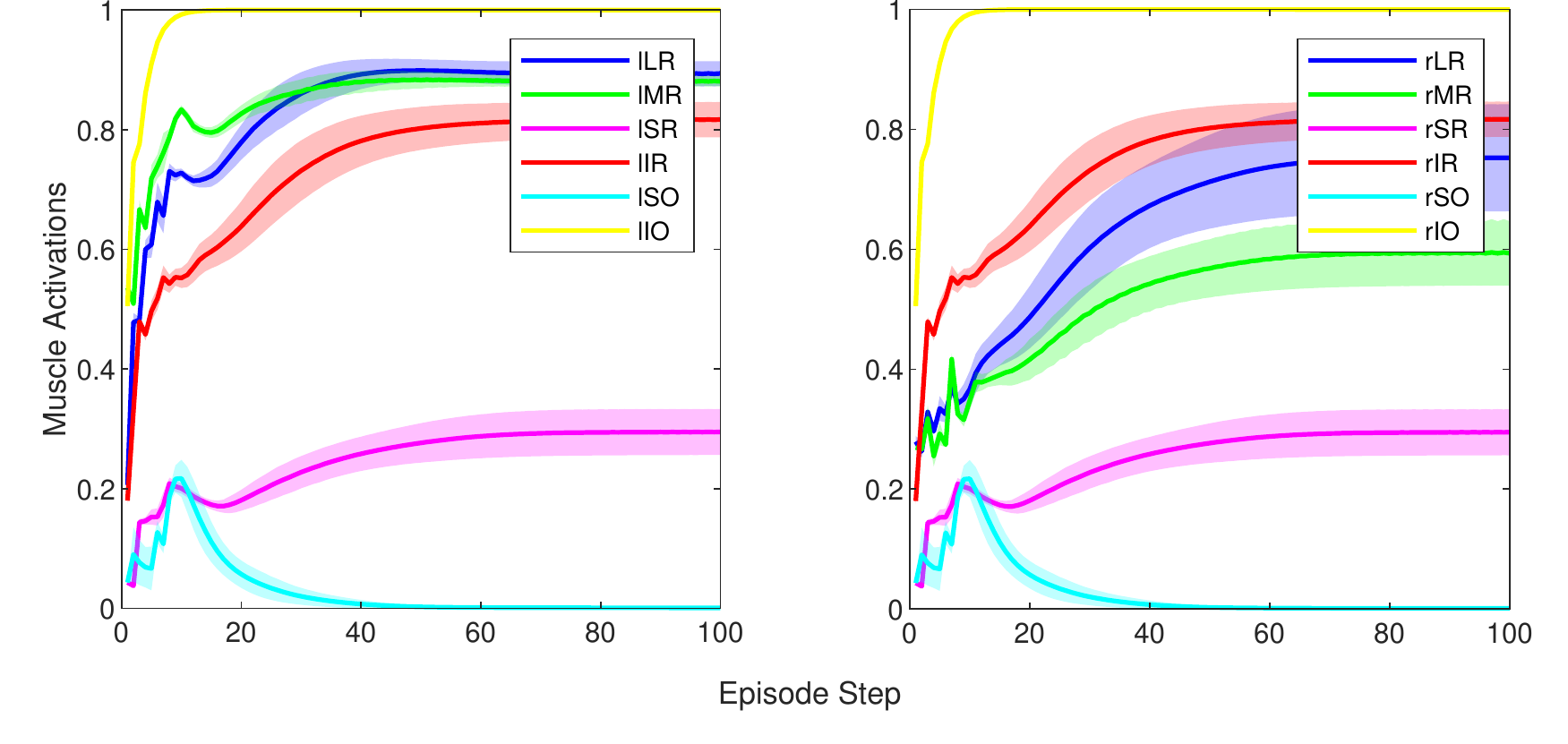}
		\label{fig:act-0.10}}
	\subfloat[(0,0)]{
		\includegraphics[trim={1.9cm 9.5cm 1.9cm 10cm},clip,width=0.33\linewidth]{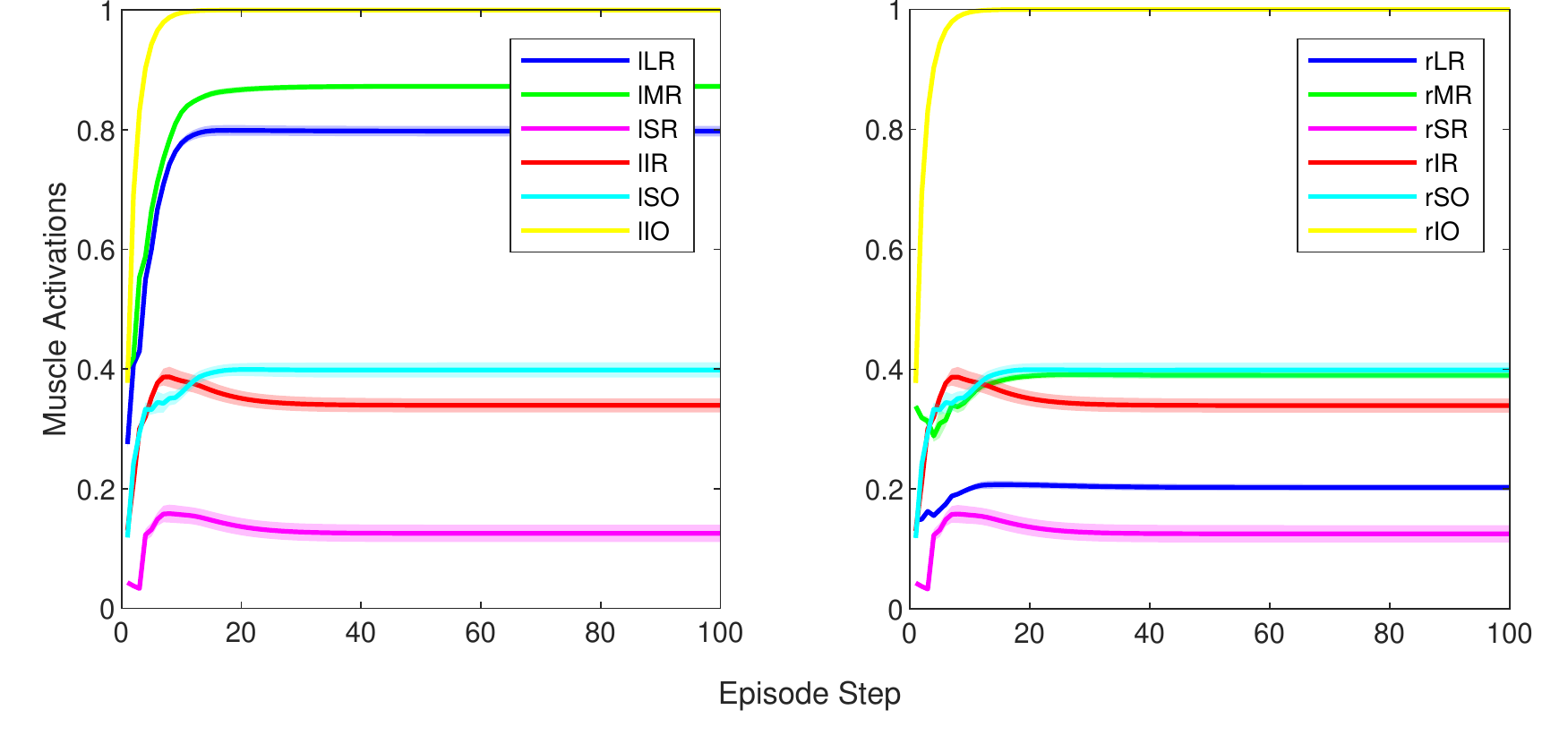}		
		\label{fig:act00}}
	\subfloat[(0.1,0.0)]{
		\includegraphics[trim={1.9cm 9.5cm 1.9cm 10cm},clip,width=0.33\linewidth]{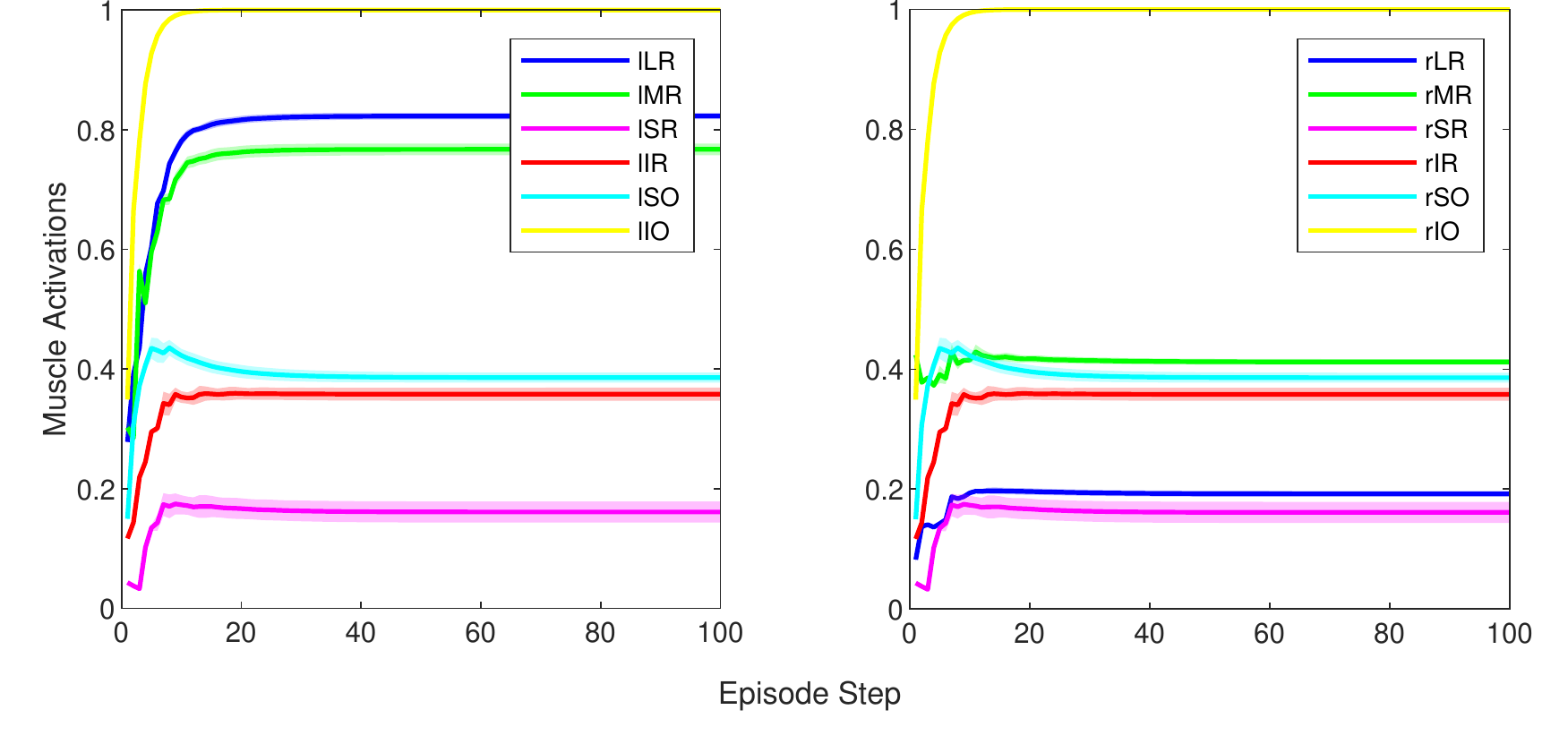}
		\label{fig:act010}}
		\\
	\subfloat[(-0.1,-0.1)]{
			\includegraphics[trim={1.9cm 9.5cm 1.9cm 10cm},clip,width=0.33\linewidth]{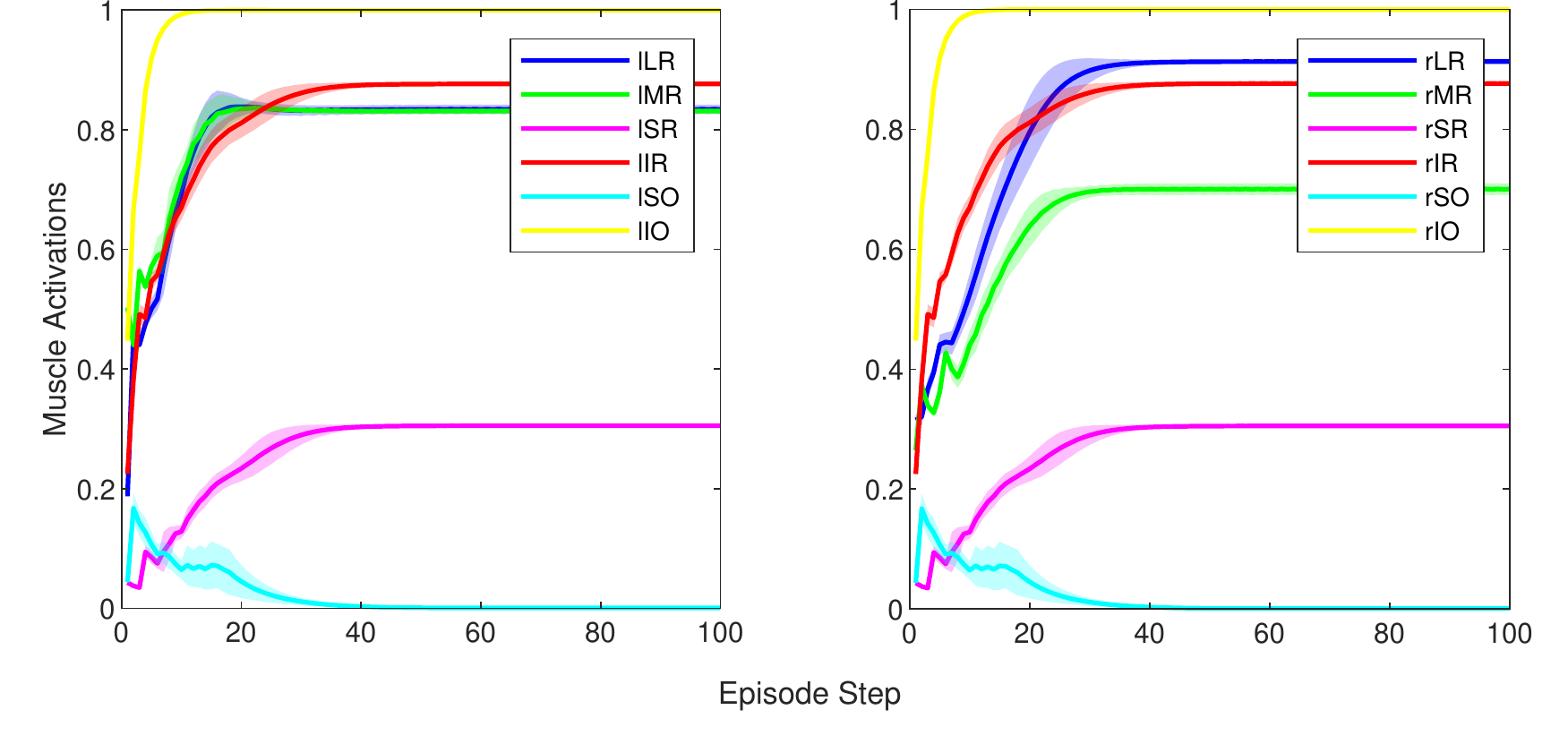}
		\label{fig:act-01-01}}
	\subfloat[(0.0,-0,1)]{
		\includegraphics[trim={1.9cm 9.5cm 1.9cm 10cm},clip,width=0.33\linewidth]{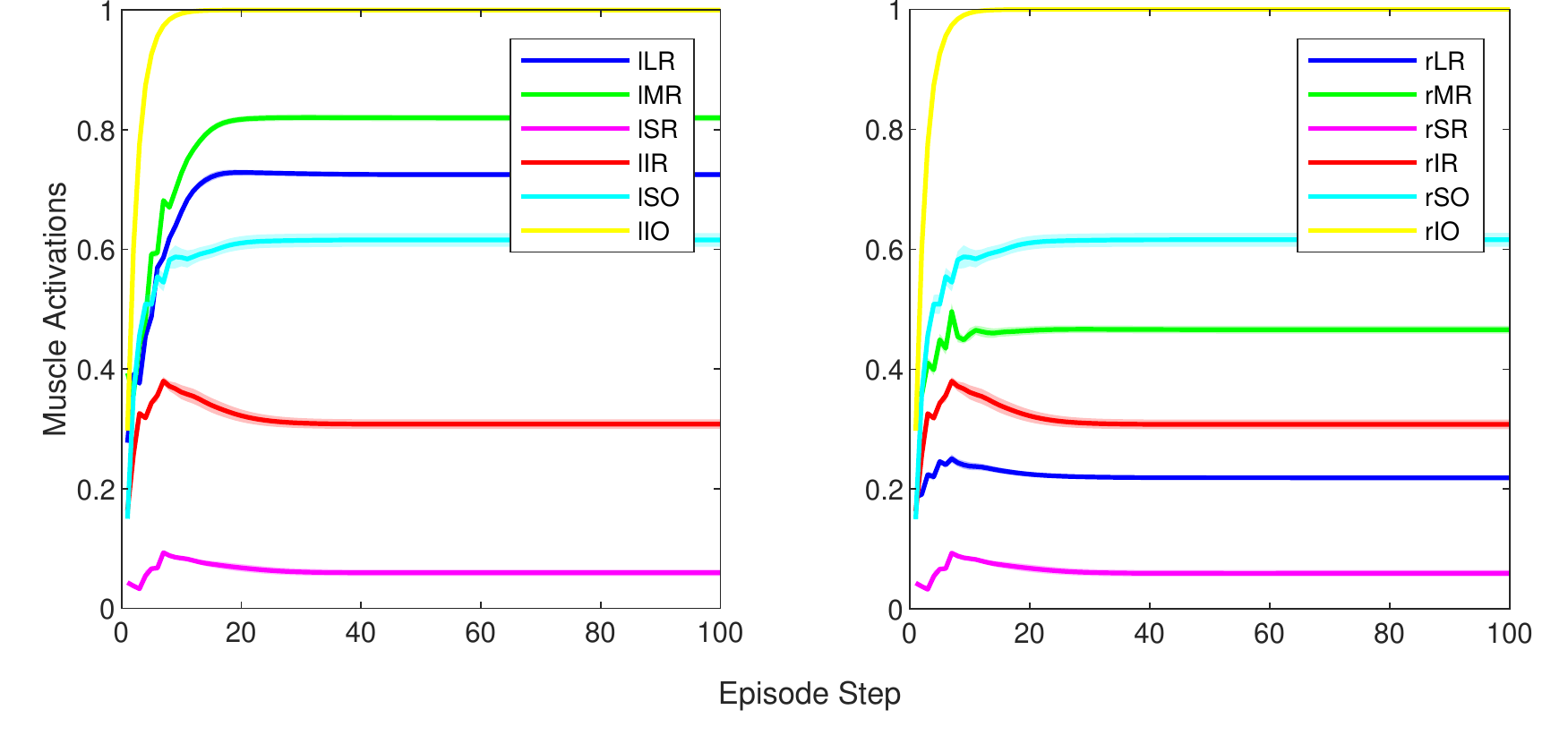}		\label{fig:act0-01}}
	\subfloat[(0.1,-0.1)]{
		\includegraphics[trim={1.9cm 9.5cm 1.9cm 10cm},clip,width=0.33\linewidth]{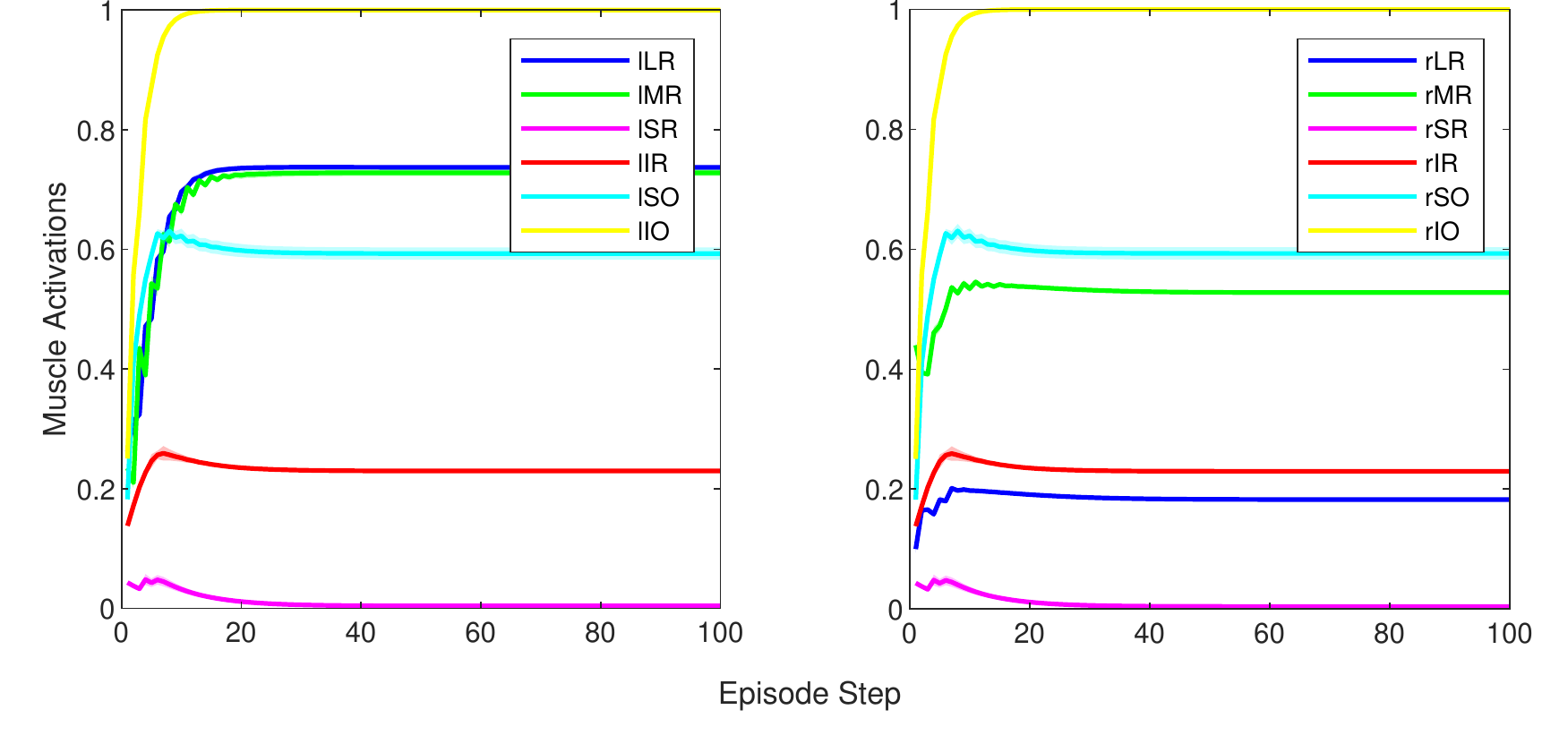}
		\label{fig:act01-01}}
		\\
\caption{Muscle activations produced from the second testing phase. Each figure represents activation produced to maintain gaze around an object that was displaced by (a) (-0.1,0.1), (b) (0.0,0.1), (c) (0.1,0.1), (d) (-0.1,0.0), (e) (0.0,0.0), (f) (0.1,0.0), (g) (-0.1,-0.1), (h) (0.0,-0.1) and (i) (0.1,-0.1) with respect to the initial position of the object. All distance measured in meters. More results in Table~\ref{tab:test2res}.}
\label{fig:testresults}
\end{figure*}

\begin{figure*}
	\centering	
	%%%trim={<left> <lower> <right> <upper>}
	\subfloat[(-0.1,0.1)]{
		\includegraphics[trim={1.9cm 5cm 1.9cm 5cm},clip,width=0.33\linewidth]{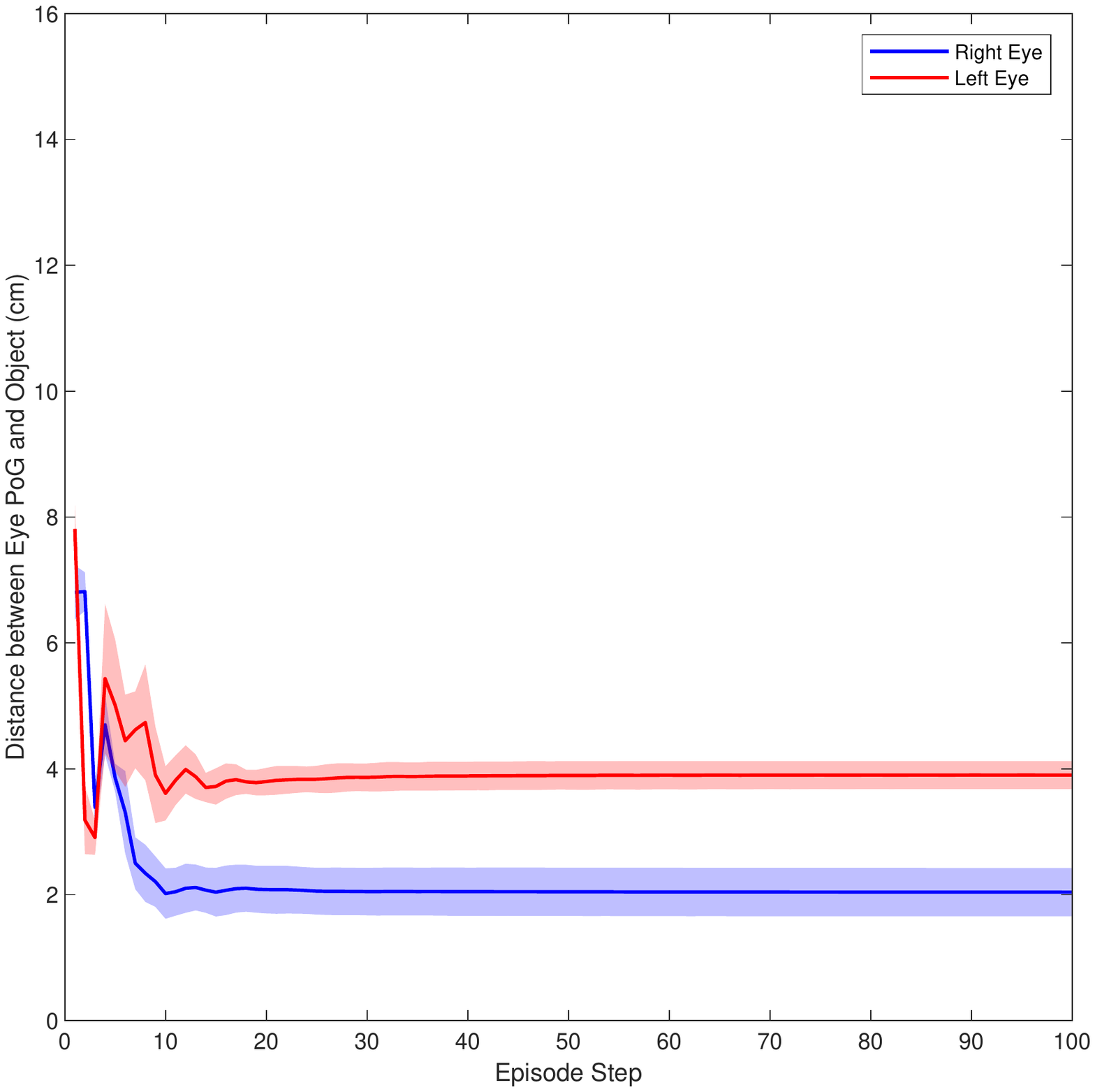}
		\label{fig:dist-0101}}
	\subfloat[(0.0,0.1)]{
		\includegraphics[trim={1.9cm 5cm 1.9cm 5cm},clip,width=0.33\linewidth]{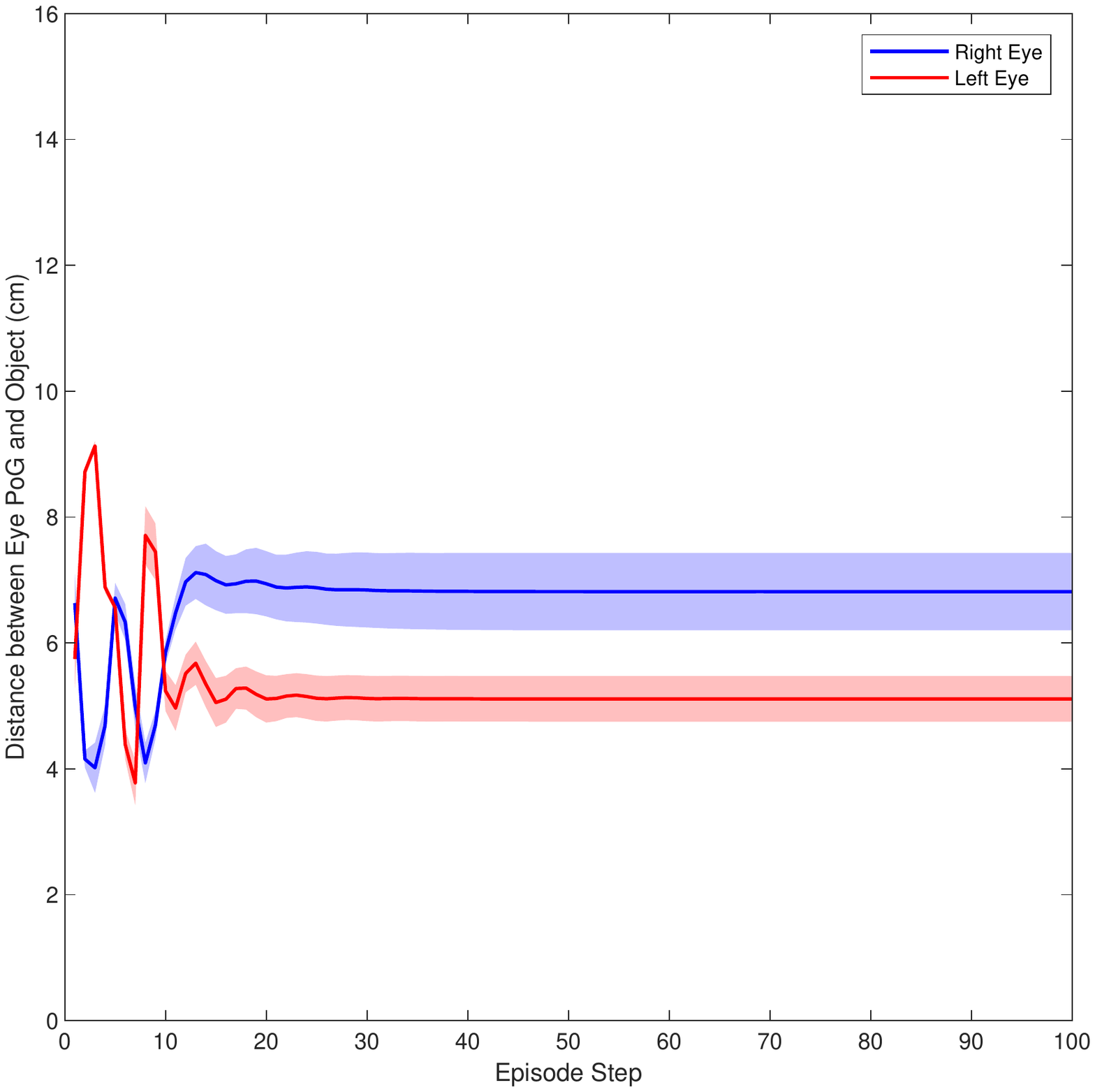}		
		\label{fig:dist00.1}}
	\subfloat[(0.1,0.1)]{
		\includegraphics[trim={1.9cm 5cm 1.9cm 5cm},clip,width=0.33\linewidth]{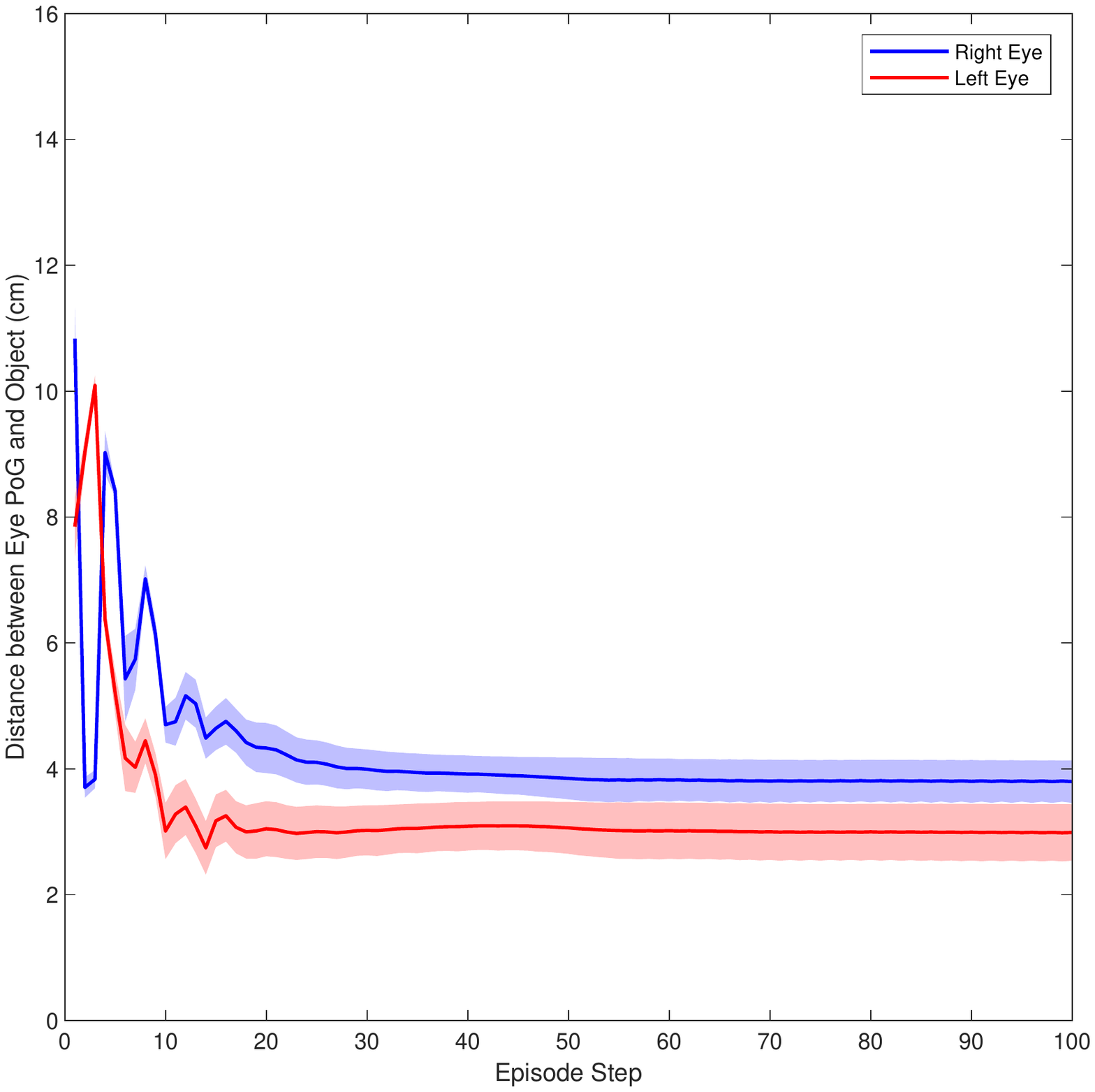}
		\label{fig:dist0.10.1}}
		\\
	\subfloat[(-0.1,0.0)]{
		\includegraphics[trim={1.9cm 5cm 1.9cm 5cm},clip,width=0.33\linewidth]{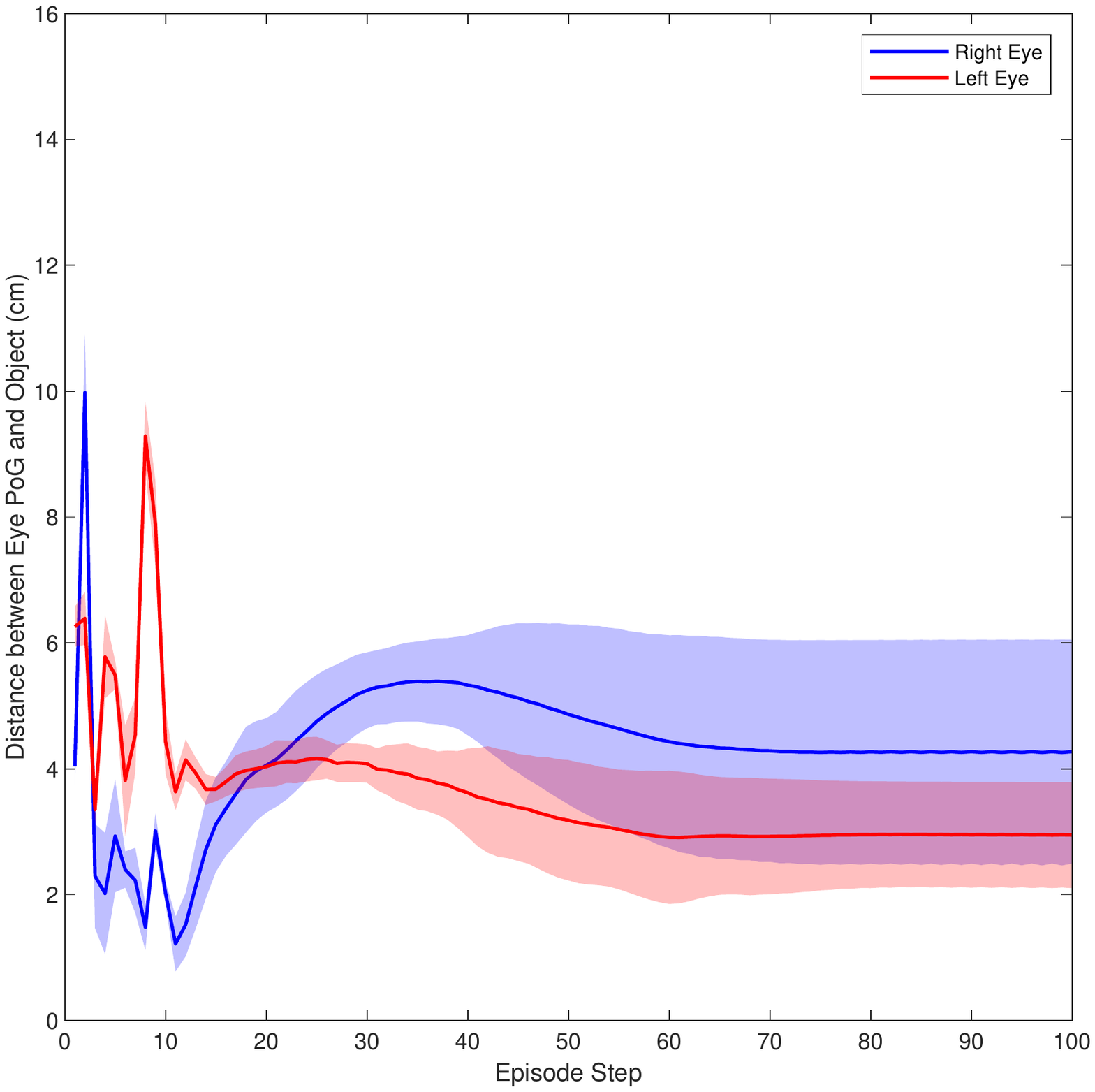}
		\label{fig:dist-0.10}}
	\subfloat[(0,0)]{
		\includegraphics[trim={1.9cm 5cm 1.9cm 5cm},clip,width=0.33\linewidth]{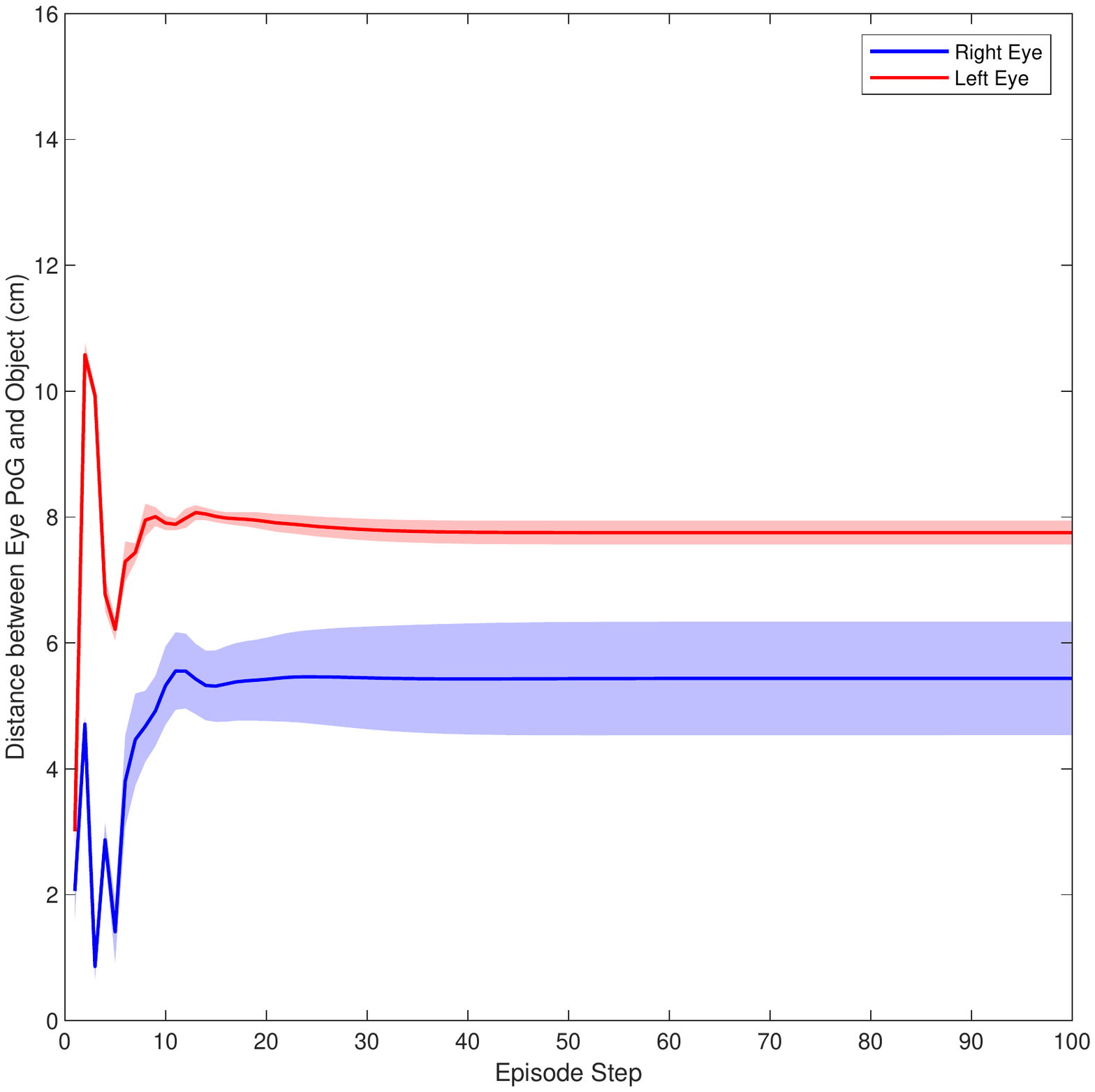}		
		\label{fig:dist00}}
	\subfloat[(0.1,0.0)]{
		\includegraphics[trim={1.9cm 5cm 1.9cm 5cm},clip,width=0.33\linewidth]{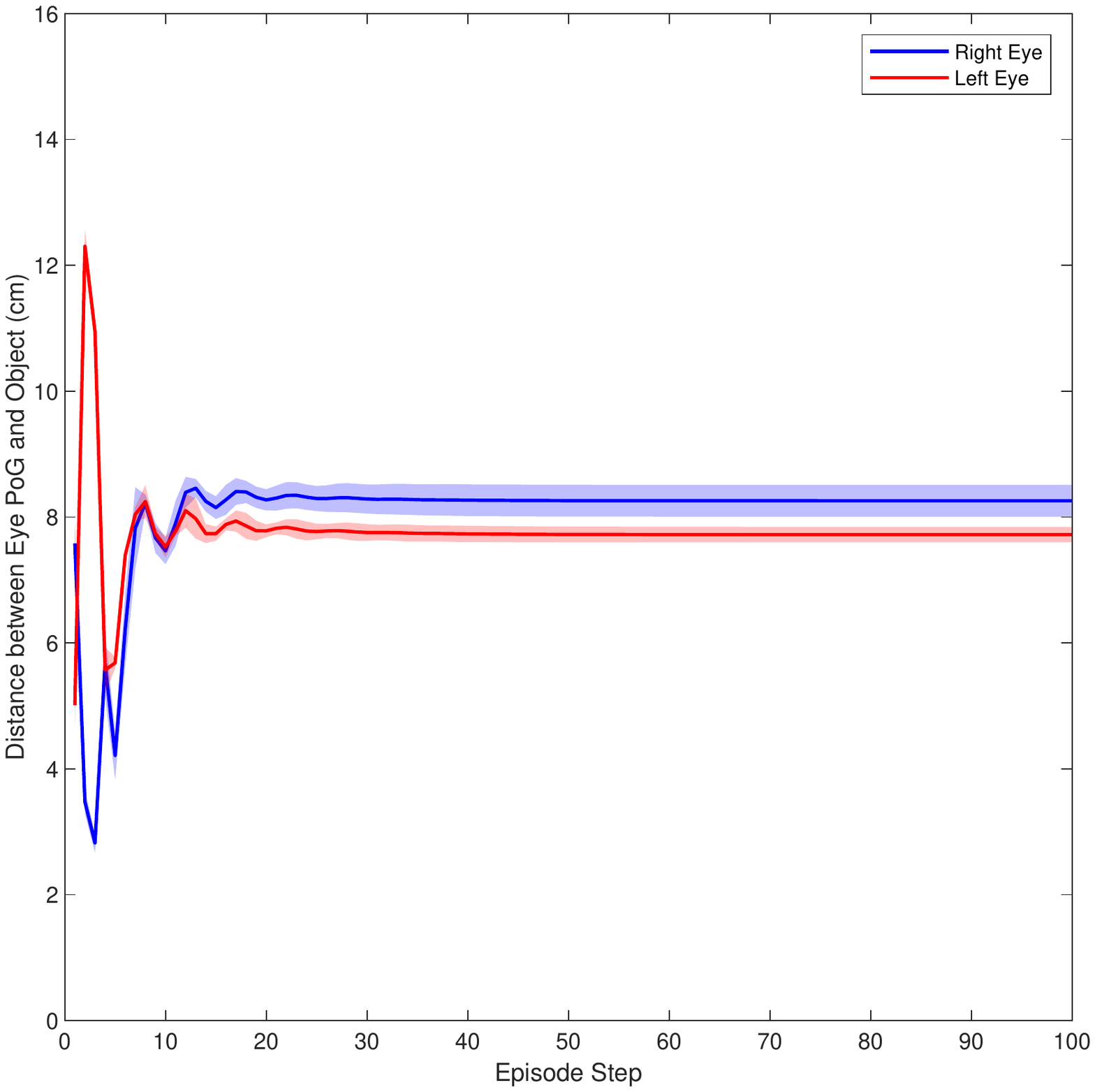}
		\label{fig:dist010}}
		\\
	\subfloat[(-0.1,-0.1)]{
		\includegraphics[trim={1.9cm 5cm 1.9cm 5cm},clip,width=0.33\linewidth]{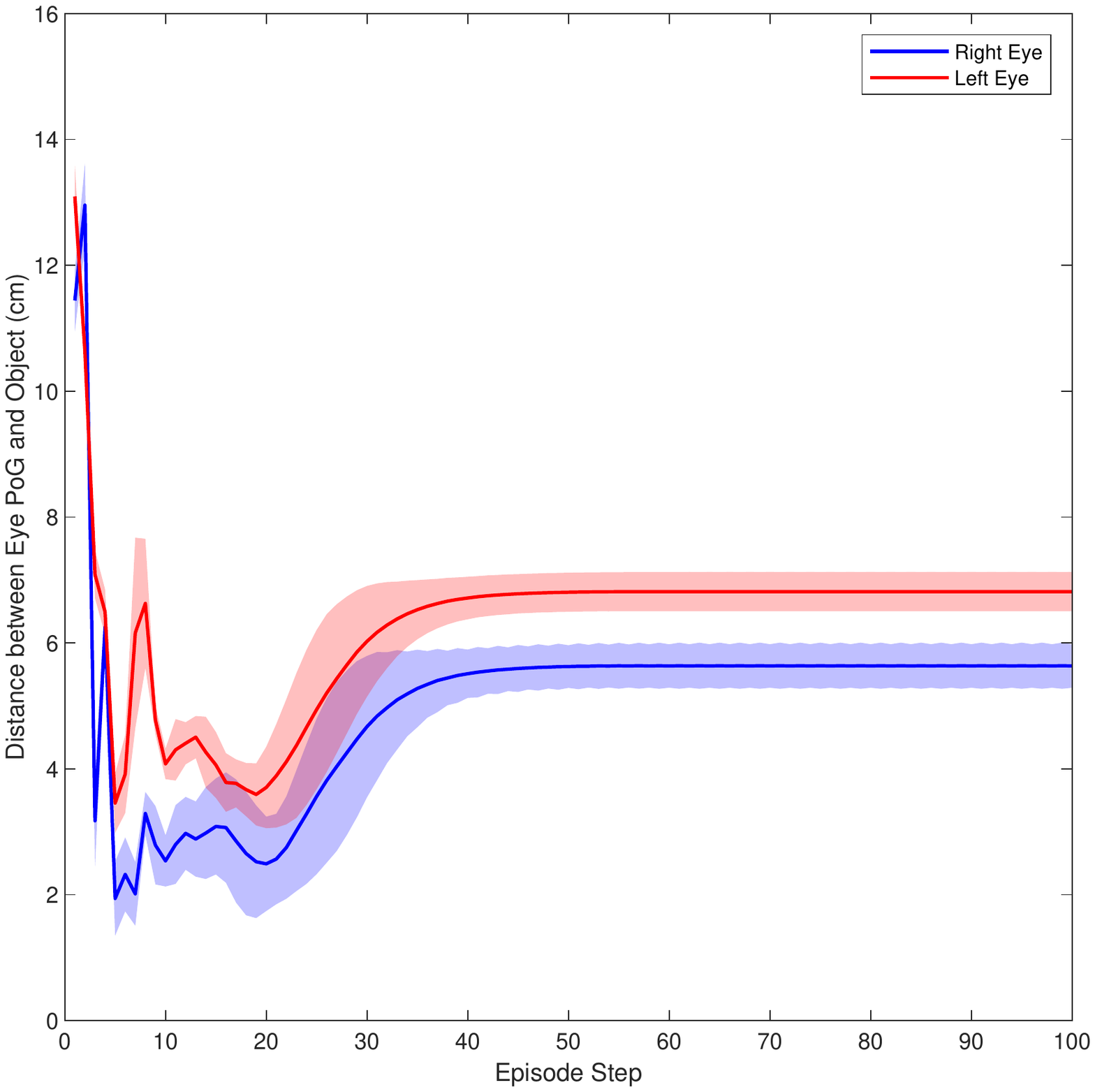}
		\label{fig:dist-01-01}}
	\subfloat[(0.0,-0,1)]{
		\includegraphics[trim={1.9cm 5cm 1.9cm 5cm},clip,width=0.33\linewidth]{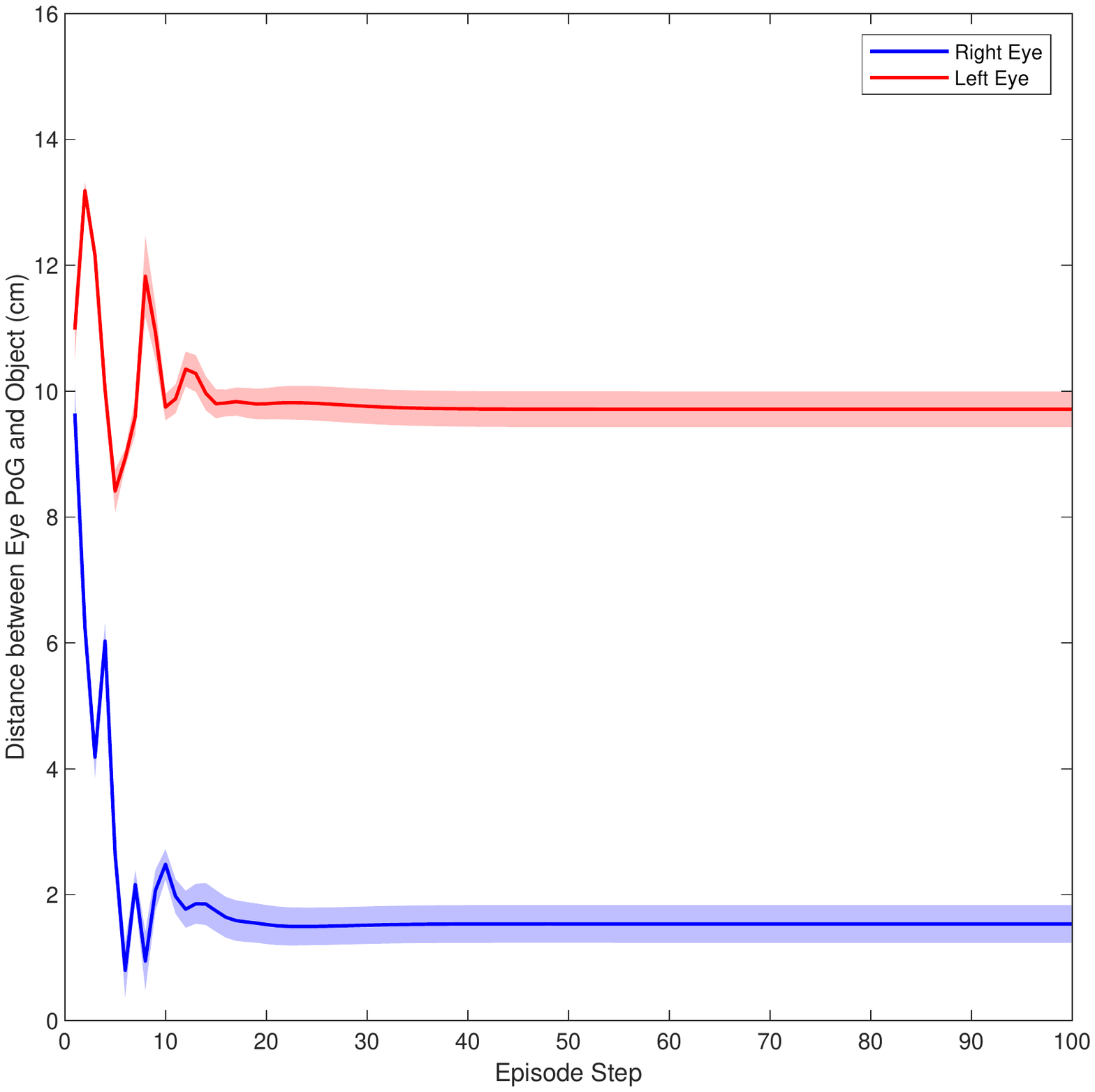}		\label{fig:dist0-01}}
	\subfloat[(0.1,-0.1)]{
		\includegraphics[trim={1.9cm 5cm 1.9cm 5cm},clip,width=0.33\linewidth]{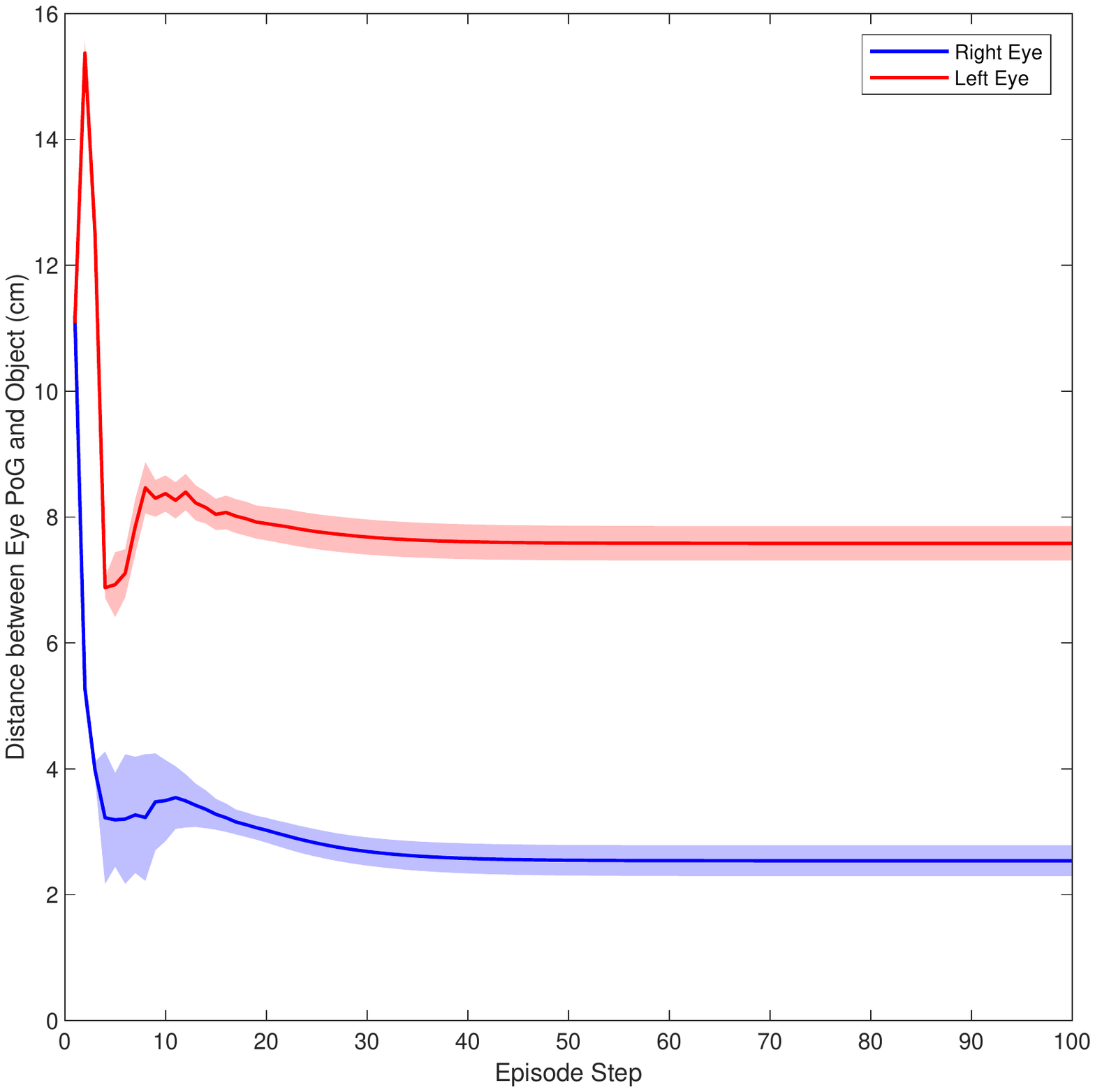}
		\label{fig:dist01-01}}
		\\
\caption{Distance between the right/left eye PoG and the object of interest. Each figure represents distance produced to maintain gaze around an object that was displaced by (a) (-0.1,0.1), (b) (0.0,0.1), (c) (0.1,0.1), (d) (-0.1,0.0), (e) (0.0,0.0), (f) (0.1,0.0), (g) (-0.1,-0.1), (h) (0.0,-0.1) and (i) (0.1,-0.1) with respect to the initial position of the object. More results in Table~\ref{tab:test2res}.}
\label{fig:distresults}
\end{figure*}

	Figure~\ref{fig:testresults} shows the muscle activation signals resulting from the second testing phase, each figure shows the mean activation of each muscle and the shaded part shows the standard deviation. Table~\ref{tab:test2res} shows mean, maximum, minimum, and standard deviation of the distance between the right/left PoG and the object of interest for each object displacement case. The mean distance was approximately $6.1\pm 2.2$ cm, which is approximately equivalent to a deviation angle of $3.5\degree~\pm 1.25\degree$, respectively, since the object is 1 m away (x-direction). The statistics were calculated after removing the first 20 steps, equivalent to $20 \times 0.01=0.2$ seconds. Figure~\ref{fig:distresults} shows the change in the distance for the right and left eye over time. 
	
\begin{table*}
	\centering
	\caption{Statistics of the distance between right(R)/left(L) PoG and the object of interest center. The first 20 steps were removed. Distance in cm.}
        %\begin{tabular}[width=\textwidth]{m{1.05cm}|m{0.39cm}m{0.44cm}|m{0.4cm}m{0.4cm}|m{0.39cm}m{0.39cm}|m{0.65cm}m{0.65cm}}
        \begin{tabular}[width=\textwidth]{|l|cc|cc|cc|m{1.2cm}m{1.2cm}|}
        	\hline
    		dy,dz  & R Mean  & L Mean &  R Max  & L Max &  R Min  & L Min & R Std Deviation & L Std Deviation \\
    		\hline
    		\hline
    		 0, 0 & 5.4 & 7.7 & 7 & 8.2 & 3.6 & 7.3 & 0.8 & 0.2\\
             0, 0.1 & 8.2 & 7.7 &  8.6 & 8.1& 7.6 & 7.5 & 0.2 & 0.12\\
             0, -0.1 & 4.6 & 3.2 & 6.4 &  4.9 & 1.4 &  0.4 &  1.5 &  0.9\\
             0.1, 0 & 6.8 & 5.1 & 7.5 & 5.7 & 5.4 & 4.4 & 0.6 & 0.3\\
    
             0.1, 0.1 & 3.8 & 3 & 5.1 & 3.9 & 3.2 & 1.9 & 0.3 &  0.4\\
            
             0.1, -0.1 & 2 & 3.9 & 2.8 & 4.3 & 1.4 & 3.4 & 0.4 & 0.2\\
             -0.1, 0 &  1.5 & 9.7 & 2.2 &  10.3 & 1.2 & 9.2 & 0.3 & 0.3\\
            
             -0.1, 0.1 &  2.6 & 7.6 & 3.4 & 8.6 & 2.1 & 7 & 0.2 & 0.3\\
             
             -0.1, -0.1 & 5.3 & 6.5 & 6.3 & 7.4 & 0.7 & 2.6  &  0.9 & 0.9\\
             
             \hline
             Overall& 4.5& 6.1 &8.6&10.3&0.7&0.45&2.2 & 2.2\\
             \hline
    		
    		\hline
    	\end{tabular}
	
    \label{tab:test2res}
\end{table*}

\section{Discussion}
\label{sc:disc}
   In this paper, we presented an ocular environment that could be used in reinforcement learning. The environment was used to train a DRL agent to learn to move the eyes and fixate on a static object at different positions with a mean deviation angle, approximately $3.5\degree~\pm 1.25\degree$. It is noted that normal eyes are not stationary even at fixations but exhibits movement of small amplitude around the region of interest~\citep{leigh2015neurology}. The DRL agent exhibited a similar behaviour; this is reflected in the shaded parts, presenting standard deviation, in Fig.~\ref{fig:testresults} and Fig.~\ref{fig:distresults}.
   
   \smallbreak
   %The reward function was chosen to represent the cost of moving away from the object; i.e. negative reward, as shown in Eq.\ref{eq:rewfun}. And thus, the agent was always trying to minimise that cost. The reward was also affected by the position of the left and right eye with respect to each other. It was penalised if the left eye crossed the right eye or vice-versa; it was also penalised if one eye was vertically higher than the other. This ensured that the agent learned to move the eyes in a coordinated fashion.
   
   \smallbreak
   The DRL agent adapted a model specific neural network that worked on capturing the EOM coordinated activation, Fig. \subref*{fig:eomctrl}. The main objective of the training was to drive the eyes to stabilise at or around the object of interest. Therefore, the agent tried to make use of all the muscles to get to the objective. The only constraint imposed was the inverse relationship between the right and left horizontal muscles (LR and MR) and the similarity between the other muscles of the left and right eyes. 
   
      \smallbreak
    The presented framework has its limitations too. As discussed, the agents main target was to maximise the reward; so it made optimal use of all muscles, without consideration of minimal effort or coordination between agonist and antagonist muscles~\citep{wong2008eye}, as it was not accounted for in the reward.  The forces of the extraocular muscles are very small compared to the forces exerted by other skeletal muscles in the upper and lower limbs and consequently the metabolic cost is balanced by the benefits of having rapid and accurate fixations~\citep{iskander2018ocular,gilchrist2011saccades,leigh2015neurology,iskander2018review}.
    
    From Fig.~\ref{fig:testresults}, the agent relied on using the IO extensively which has a primary action of excyclotorsion and a secondary and tertiary action of elevation and abduction, respectively~\citep{wong2008eye}, Table~\ref{tab:EOMfunc}. In contrast, LR and MR are used solely for abduction and adduction, respectively. Therefore, to compensate the high activation of IO, the IR  was in most cases activated to compensate for the elevation caused by the IO. 
    
    From Figures \ref{fig:results} and \ref{fig:testresults}, we can see that the agent was aspiring to achieve the best results; and it used most of the muscles all the time to achieve that objective. Although this may not be physiologically sound, it is a first step towards training a DRL agent for ocular motility. The next stage of this research will focus on fine tuning the actor and critic neural networks to achieve better results comparable to the ocular control theories established~\citep{robinson1975basic,jurgens1981natural,scudder2002brainstem,sparks2002brainstem}. The environment can be also modified to simulate neck movement and thus, simulate vestibulo-ocular reflex and also, eye-hand coordination. Numerous scenarios can be simulated, from normal to pathological scenarios, and then analysis can be followed which will highlight how different control strategies can be used to perform improved eye movements. The DRL agent training can also be aligned with paediatrics research in ocular development.

\section*{Conflict of Interest Declaration}
None

\bibliographystyle{elsarticle-harv}
%\bibliography{bibs/julie_isk_my,bibs/julie_isk_eyebiomech,bibs/julie_isk_RL,bibs/julie_isk_biomech}

\begin{thebibliography}{30}
\expandafter\ifx\csname natexlab\endcsname\relax\def\natexlab#1{#1}\fi
\providecommand{\url}[1]{\texttt{#1}}
\providecommand{\href}[2]{#2}
\providecommand{\path}[1]{#1}
\providecommand{\DOIprefix}{doi:}
\providecommand{\ArXivprefix}{arXiv:}
\providecommand{\URLprefix}{URL: }
\providecommand{\Pubmedprefix}{pmid:}
\providecommand{\doi}[1]{\href{http://dx.doi.org/#1}{\path{#1}}}
\providecommand{\Pubmed}[1]{\href{pmid:#1}{\path{#1}}}
\providecommand{\bibinfo}[2]{#2}
\ifx\xfnm\relax \def\xfnm[#1]{\unskip,\space#1}\fi
%Type = Article
\bibitem[{Brockman et~al.(2016)Brockman, Cheung, Pettersson, Schneider,
  Schulman, Tang and Zaremba}]{brockman2016openai}
\bibinfo{author}{Brockman, G.}, \bibinfo{author}{Cheung, V.},
  \bibinfo{author}{Pettersson, L.}, \bibinfo{author}{Schneider, J.},
  \bibinfo{author}{Schulman, J.}, \bibinfo{author}{Tang, J.},
  \bibinfo{author}{Zaremba, W.}, \bibinfo{year}{2016}.
\newblock \bibinfo{title}{{OpenAI Gym}}.
\newblock \bibinfo{journal}{arXiv preprint arXiv:1606.01540} .
%Type = Incollection
\bibitem[{Gilchrist(2011)}]{gilchrist2011saccades}
\bibinfo{author}{Gilchrist, I.}, \bibinfo{year}{2011}.
\newblock \bibinfo{title}{Saccades}, in: \bibinfo{booktitle}{The Oxford
  handbook of eye movements}.
%Type = Book
\bibitem[{Hering et~al.(1977)Hering, Bridgeman and Stark}]{hering1977theory}
\bibinfo{author}{Hering, E.}, \bibinfo{author}{Bridgeman, B.},
  \bibinfo{author}{Stark, L.}, \bibinfo{year}{1977}.
\newblock \bibinfo{title}{The theory of binocular vision}.
\newblock \bibinfo{publisher}{Springer}.
%Type = Article
\bibitem[{Hossny and Iskander(2020)}]{HossnyIskander2020_dontfall}
\bibinfo{author}{Hossny, M.}, \bibinfo{author}{Iskander, J.},
  \bibinfo{year}{2020}.
\newblock \bibinfo{title}{Just don't fall: An ai agent's learning journey
  towards posture stabilisation}.
\newblock \bibinfo{journal}{AI} \bibinfo{volume}{1}, \bibinfo{pages}{286--298.}
%Type = Article
\bibitem[{Hossny et~al.(2020)Hossny, Iskander, Attia and
  Saleh}]{HossnyEtal2020_PTANH}
\bibinfo{author}{Hossny, M.}, \bibinfo{author}{Iskander, J.},
  \bibinfo{author}{Attia, M.}, \bibinfo{author}{Saleh, K.},
  \bibinfo{year}{2020}.
\newblock \bibinfo{title}{Refined continuous control of ddpg actors via
  parametrised activation}.
\newblock \bibinfo{journal}{arXiv:2006.02818 [cs.LG]}
  \href{http://arxiv.org/abs/2006.02818}{{\tt arXiv:2006.02818}}.
%Type = Inproceedings
\bibitem[{Iskander et~al.(2018a)Iskander, Hanoun, Hettiarachchi, Hossny, Saleh,
  Zhou, Nahavandi and Bhatti}]{iskander2018eye}
\bibinfo{author}{Iskander, J.}, \bibinfo{author}{Hanoun, S.},
  \bibinfo{author}{Hettiarachchi, I.}, \bibinfo{author}{Hossny, M.},
  \bibinfo{author}{Saleh, K.}, \bibinfo{author}{Zhou, H.},
  \bibinfo{author}{Nahavandi, S.}, \bibinfo{author}{Bhatti, A.},
  \bibinfo{year}{2018}a.
\newblock \bibinfo{title}{Eye behaviour as a hazard perception measure}, in:
  \bibinfo{booktitle}{Systems Conference (SysCon), 2018 Annual IEEE
  International}, \bibinfo{organization}{IEEE}. pp. \bibinfo{pages}{1--6}.
%Type = Inproceedings
\bibitem[{Iskander et~al.(2018b)Iskander, Hossny and
  Nahavandi}]{iskander2018biomechanical}
\bibinfo{author}{Iskander, J.}, \bibinfo{author}{Hossny, M.},
  \bibinfo{author}{Nahavandi, S.}, \bibinfo{year}{2018}b.
\newblock \bibinfo{title}{Biomechanical analysis of eye movement in virtual
  environments: A validation study}, in: \bibinfo{booktitle}{2018 IEEE
  International Conference on Systems, Man, and Cybernetics (SMC)},
  \bibinfo{organization}{IEEE}. pp. \bibinfo{pages}{3498--3503}.
%Type = Article
\bibitem[{Iskander et~al.(2018c)Iskander, Hossny and
  Nahavandi}]{iskander2018review}
\bibinfo{author}{Iskander, J.}, \bibinfo{author}{Hossny, M.},
  \bibinfo{author}{Nahavandi, S.}, \bibinfo{year}{2018}c.
\newblock \bibinfo{title}{A review on ocular biomechanic models for assessing
  visual fatigue in virtual reality}.
\newblock \bibinfo{journal}{IEEE Access} \bibinfo{volume}{6},
  \bibinfo{pages}{19345--19361}.
\newblock \DOIprefix\doi{10.1109/ACCESS.2018.2815663}.
%Type = Article
\bibitem[{Iskander et~al.(2019)Iskander, Hossny and
  Nahavandi}]{iskander2019using}
\bibinfo{author}{Iskander, J.}, \bibinfo{author}{Hossny, M.},
  \bibinfo{author}{Nahavandi, S.}, \bibinfo{year}{2019}.
\newblock \bibinfo{title}{{Using biomechanics to investigate the effect of VR
  on eye vergence system}}.
\newblock \bibinfo{journal}{Applied Ergonomics} \bibinfo{volume}{81},
  \bibinfo{pages}{102883}.
\newblock \DOIprefix\doi{10.1016/j.apergo.2019.102883}.
%Type = Article
\bibitem[{Iskander et~al.(2018d)Iskander, Hossny, Nahavandi and
  Del~Porto}]{iskander2018ocular}
\bibinfo{author}{Iskander, J.}, \bibinfo{author}{Hossny, M.},
  \bibinfo{author}{Nahavandi, S.}, \bibinfo{author}{Del~Porto, L.},
  \bibinfo{year}{2018}d.
\newblock \bibinfo{title}{An ocular biomechanic model for dynamic simulation of
  different eye movements}.
\newblock \bibinfo{journal}{Journal of biomechanics} \bibinfo{volume}{71},
  \bibinfo{pages}{208--216}.
%Type = Article
\bibitem[{J{\"u}rgens et~al.(1981)J{\"u}rgens, Becker and
  Kornhuber}]{jurgens1981natural}
\bibinfo{author}{J{\"u}rgens, R.}, \bibinfo{author}{Becker, W.},
  \bibinfo{author}{Kornhuber, H.}, \bibinfo{year}{1981}.
\newblock \bibinfo{title}{Natural and drug-induced variations of velocity and
  duration of human saccadic eye movements: evidence for a control of the
  neural pulse generator by local feedback}.
\newblock \bibinfo{journal}{Biological cybernetics} \bibinfo{volume}{39},
  \bibinfo{pages}{87--96}.
%Type = Incollection
\bibitem[{Kidzi{\'n}ski et~al.(2018a)Kidzi{\'n}ski, Mohanty, Ong, Hicks,
  Carroll, Levine, Salath{\'e} and Delp}]{kidzinski2018learning}
\bibinfo{author}{Kidzi{\'n}ski, {\L}.}, \bibinfo{author}{Mohanty, S.P.},
  \bibinfo{author}{Ong, C.F.}, \bibinfo{author}{Hicks, J.L.},
  \bibinfo{author}{Carroll, S.F.}, \bibinfo{author}{Levine, S.},
  \bibinfo{author}{Salath{\'e}, M.}, \bibinfo{author}{Delp, S.L.},
  \bibinfo{year}{2018}a.
\newblock \bibinfo{title}{Learning to run challenge: Synthesizing
  physiologically accurate motion using deep reinforcement learning}, in:
  \bibinfo{booktitle}{The NIPS'17 Competition: Building Intelligent Systems}.
  \bibinfo{publisher}{Springer}, pp. \bibinfo{pages}{101--120}.
%Type = Incollection
\bibitem[{Kidzi{\'n}ski et~al.(2018b)Kidzi{\'n}ski, Mohanty, Ong, Huang, Zhou,
  Pechenko, Stelmaszczyk, Jarosik, Pavlov, Kolesnikov
  et~al.}]{kidzinski2018learningsolu}
\bibinfo{author}{Kidzi{\'n}ski, {\L}.}, \bibinfo{author}{Mohanty, S.P.},
  \bibinfo{author}{Ong, C.F.}, \bibinfo{author}{Huang, Z.},
  \bibinfo{author}{Zhou, S.}, \bibinfo{author}{Pechenko, A.},
  \bibinfo{author}{Stelmaszczyk, A.}, \bibinfo{author}{Jarosik, P.},
  \bibinfo{author}{Pavlov, M.}, \bibinfo{author}{Kolesnikov, S.}, et~al.,
  \bibinfo{year}{2018}b.
\newblock \bibinfo{title}{Learning to run challenge solutions: Adapting
  reinforcement learning methods for neuromusculoskeletal environments}, in:
  \bibinfo{booktitle}{The NIPS'17 Competition: Building Intelligent Systems}.
  \bibinfo{publisher}{Springer}, pp. \bibinfo{pages}{121--153}.
%Type = Incollection
\bibitem[{Kidzi{\'n}ski et~al.(2020)Kidzi{\'n}ski, Ong, Mohanty, Hicks,
  Carroll, Zhou, Zeng, Wang, Lian, Tian et~al.}]{kidzinski2020artificial}
\bibinfo{author}{Kidzi{\'n}ski, {\L}.}, \bibinfo{author}{Ong, C.},
  \bibinfo{author}{Mohanty, S.P.}, \bibinfo{author}{Hicks, J.},
  \bibinfo{author}{Carroll, S.}, \bibinfo{author}{Zhou, B.},
  \bibinfo{author}{Zeng, H.}, \bibinfo{author}{Wang, F.},
  \bibinfo{author}{Lian, R.}, \bibinfo{author}{Tian, H.}, et~al.,
  \bibinfo{year}{2020}.
\newblock \bibinfo{title}{Artificial intelligence for prosthetics: Challenge
  solutions}, in: \bibinfo{booktitle}{The NeurIPS'18 Competition}.
  \bibinfo{publisher}{Springer}, pp. \bibinfo{pages}{69--128}.
%Type = Article
\bibitem[{Kingma and Ba(2014)}]{kingma2014adam}
\bibinfo{author}{Kingma, D.P.}, \bibinfo{author}{Ba, J.}, \bibinfo{year}{2014}.
\newblock \bibinfo{title}{Adam: A method for stochastic optimization}.
\newblock \bibinfo{journal}{arXiv preprint arXiv:1412.6980} .
%Type = Inproceedings
\bibitem[{Konda and Tsitsiklis(2000)}]{konda2000actor}
\bibinfo{author}{Konda, V.R.}, \bibinfo{author}{Tsitsiklis, J.N.},
  \bibinfo{year}{2000}.
\newblock \bibinfo{title}{Actor-critic algorithms}, in:
  \bibinfo{booktitle}{Advances in neural information processing systems}, pp.
  \bibinfo{pages}{1008--1014}.
%Type = Book
\bibitem[{Leigh and Zee(2015)}]{leigh2015neurology}
\bibinfo{author}{Leigh, R.J.}, \bibinfo{author}{Zee, D.S.},
  \bibinfo{year}{2015}.
\newblock \bibinfo{title}{The neurology of eye movements}.
  volume~\bibinfo{volume}{90}.
\newblock \bibinfo{publisher}{Oxford University Press, USA}.
%Type = Article
\bibitem[{Lillicrap et~al.(2015)Lillicrap, Hunt, Pritzel, Heess, Erez, Tassa,
  Silver and Wierstra}]{lillicrap2015continuous}
\bibinfo{author}{Lillicrap, T.P.}, \bibinfo{author}{Hunt, J.J.},
  \bibinfo{author}{Pritzel, A.}, \bibinfo{author}{Heess, N.},
  \bibinfo{author}{Erez, T.}, \bibinfo{author}{Tassa, Y.},
  \bibinfo{author}{Silver, D.}, \bibinfo{author}{Wierstra, D.},
  \bibinfo{year}{2015}.
\newblock \bibinfo{title}{Continuous control with deep reinforcement learning}.
\newblock \bibinfo{journal}{arXiv preprint arXiv:1509.02971} .
%Type = Article
\bibitem[{Millard et~al.(2013)Millard, Uchida, Seth and
  Delp}]{millard2013flexing}
\bibinfo{author}{Millard, M.}, \bibinfo{author}{Uchida, T.},
  \bibinfo{author}{Seth, A.}, \bibinfo{author}{Delp, S.L.},
  \bibinfo{year}{2013}.
\newblock \bibinfo{title}{Flexing computational muscle: modeling and simulation
  of musculotendon dynamics}.
\newblock \bibinfo{journal}{Journal of biomechanical engineering}
  \bibinfo{volume}{135}, \bibinfo{pages}{021005}.
%Type = Article
\bibitem[{Purves et~al.(2001)Purves, Augustine, Fitzpatrick, Katz, LaMantia,
  McNamara and Williams}]{purves2001neural}
\bibinfo{author}{Purves, D.}, \bibinfo{author}{Augustine, G.},
  \bibinfo{author}{Fitzpatrick, D.}, \bibinfo{author}{Katz, L.},
  \bibinfo{author}{LaMantia, A.}, \bibinfo{author}{McNamara, J.},
  \bibinfo{author}{Williams, S.}, \bibinfo{year}{2001}.
\newblock \bibinfo{title}{Neural control of saccadic eye movements}.
\newblock \bibinfo{journal}{Neuroscience. Sutherland (MA): Sinauer Associates}
  .
%Type = Inbook
\bibitem[{Robinson et~al.(1975)Robinson, Lennerstrand and Bach-y
  Rita}]{robinson1975basic}
\bibinfo{author}{Robinson, D.}, \bibinfo{author}{Lennerstrand, G.},
  \bibinfo{author}{Bach-y Rita, P.}, \bibinfo{year}{1975}.
\newblock \bibinfo{title}{Basic mechanisms of ocular motility and their
  clinical implications}. \bibinfo{publisher}{Pergamon}.
  volume~\bibinfo{volume}{24}.
%Type = Article
\bibitem[{Scudder et~al.(2002)Scudder, Kaneko and Fuchs}]{scudder2002brainstem}
\bibinfo{author}{Scudder, C.A.}, \bibinfo{author}{Kaneko, C.R.},
  \bibinfo{author}{Fuchs, A.F.}, \bibinfo{year}{2002}.
\newblock \bibinfo{title}{The brainstem burst generator for saccadic eye
  movements}.
\newblock \bibinfo{journal}{Experimental brain research} \bibinfo{volume}{142},
  \bibinfo{pages}{439--462}.
%Type = Article
\bibitem[{Seth et~al.(2018)Seth, Hicks, Uchida, Habib, Dembia, Dunne, Ong,
  DeMers, Rajagopal, Millard et~al.}]{seth2018opensim}
\bibinfo{author}{Seth, A.}, \bibinfo{author}{Hicks, J.L.},
  \bibinfo{author}{Uchida, T.K.}, \bibinfo{author}{Habib, A.},
  \bibinfo{author}{Dembia, C.L.}, \bibinfo{author}{Dunne, J.J.},
  \bibinfo{author}{Ong, C.F.}, \bibinfo{author}{DeMers, M.S.},
  \bibinfo{author}{Rajagopal, A.}, \bibinfo{author}{Millard, M.}, et~al.,
  \bibinfo{year}{2018}.
\newblock \bibinfo{title}{Opensim: Simulating musculoskeletal dynamics and
  neuromuscular control to study human and animal movement}.
\newblock \bibinfo{journal}{PLoS computational biology} \bibinfo{volume}{14},
  \bibinfo{pages}{e1006223}.
%Type = Article
\bibitem[{Sherrington(1893)}]{sherrington1893ii}
\bibinfo{author}{Sherrington, C.S.}, \bibinfo{year}{1893}.
\newblock \bibinfo{title}{Ii. note on the knee-jerk and the correlation of
  action of antagonistic muscles}.
\newblock \bibinfo{journal}{Proceedings of the Royal Society of London}
  \bibinfo{volume}{52}, \bibinfo{pages}{556--564}.
%Type = Inproceedings
\bibitem[{Silver et~al.(2014)Silver, Lever, Heess, Degris, Wierstra and
  Riedmiller}]{silver2014deterministic}
\bibinfo{author}{Silver, D.}, \bibinfo{author}{Lever, G.},
  \bibinfo{author}{Heess, N.}, \bibinfo{author}{Degris, T.},
  \bibinfo{author}{Wierstra, D.}, \bibinfo{author}{Riedmiller, M.},
  \bibinfo{year}{2014}.
\newblock \bibinfo{title}{Deterministic policy gradient algorithms}, in:
  \bibinfo{booktitle}{International Conference on Machine Learning}, pp.
  \bibinfo{pages}{387--395}.
%Type = Article
\bibitem[{Sparks(2002)}]{sparks2002brainstem}
\bibinfo{author}{Sparks, D.L.}, \bibinfo{year}{2002}.
\newblock \bibinfo{title}{The brainstem control of saccadic eye movements}.
\newblock \bibinfo{journal}{Nature Reviews Neuroscience} \bibinfo{volume}{3},
  \bibinfo{pages}{952--964}.
%Type = Book
\bibitem[{Sutton and Barto(2018)}]{sutton2018introduction}
\bibinfo{author}{Sutton, R.S.}, \bibinfo{author}{Barto, A.G.},
  \bibinfo{year}{2018}.
\newblock \bibinfo{title}{Introduction to reinforcement learning, 2nd ed.}
\newblock \bibinfo{publisher}{MIT press Cambridge}.
%Type = Article
\bibitem[{Thelen(2003)}]{thelen2003adjustment}
\bibinfo{author}{Thelen, D.G.}, \bibinfo{year}{2003}.
\newblock \bibinfo{title}{Adjustment of muscle mechanics model parameters to
  simulate dynamic contractions in older adults}.
\newblock \bibinfo{journal}{Journal of biomechanical engineering}
  \bibinfo{volume}{125}, \bibinfo{pages}{70--77}.
%Type = Book
\bibitem[{Von~Noorden and Campos(2002)}]{von2002binocular}
\bibinfo{author}{Von~Noorden, G.K.}, \bibinfo{author}{Campos, E.C.},
  \bibinfo{year}{2002}.
\newblock \bibinfo{title}{Binocular vision and ocular motility}.
\newblock \bibinfo{publisher}{Mosby}.
%Type = Book
\bibitem[{Wong(2008)}]{wong2008eye}
\bibinfo{author}{Wong, A.M.F.}, \bibinfo{year}{2008}.
\newblock \bibinfo{title}{Eye movement disorders}.
\newblock \bibinfo{publisher}{Oxford University Press}.

\end{thebibliography}

\end{document}